\documentclass[manuscript]{acmart}
\AtBeginDocument{%
  \providecommand\BibTeX{{%
    \normalfont B\kern-0.5em{\scshape i\kern-0.25em b}\kern-0.8em\TeX}}}

\setcopyright{acmcopyright}
\copyrightyear{2023}
\acmYear{2023}
\acmDOI{XXXXXXX.XXXXXXX}

\acmJournal{JACM}
\acmVolume{37}
\acmNumber{4}
\acmArticle{111}
\acmMonth{8}
  
\acmPrice{15.00}
\acmISBN{978-1-4503-XXXX-X/18/06}

\newcommand{\ie}{\textit{i.e., }}
\newcommand{\etal}{\textit{et al. }}
\newcommand{\eg}{\textit{e.g., }}
\newcommand{\etc}{\textit{etc}}
\usepackage{pifont}% http://ctan.org/pkg/pifont
\newcommand{\cmark}{\ding{51}}%
\usepackage{wrapfig}
\usepackage{multirow}
\usepackage{adjustbox}
\usepackage{tikz}
\usetikzlibrary{mindmap, shadows}
\usepackage{subfigure}
\usepackage{hyperref}

\begin{document}

\title{A Survey of Dataset Refinement for Problems in Computer Vision Datasets}

\author{Zhijing Wan}
\orcid{0000-0002-1273-9043}
\affiliation{%
  \institution{National Engineering Research Center for Multimedia Software, Institute of Artificial Intelligence, School of Computer Science, Wuhan University}
  \city{Wuhan}
  \country{China}}
\email{wanzjwhu@whu.edu.cn}

\author{Zhixiang Wang}
\orcid{0000-0002-5016-587X}
\affiliation{%
  \institution{Graduate School of Information Science and Technology, The University of Tokyo}
  \city{Tokyo}
  \country{Japan}}
\email{wangzx1994@gmail.com}

\author{CheukTing Chung}
\orcid{0009-0009-7260-4077}
\affiliation{%
  \institution{National Engineering Research Center for Multimedia Software, Institute of Artificial Intelligence, School of Computer Science, Wuhan University}
  \city{Wuhan}
  \country{China}}
\email{2271406579@qq.com}

\author{Zheng Wang}
\orcid{0000-0003-3846-9157}
\authornotemark[2]
\renewcommand{\thefootnote}{\fnsymbol{footnote}}
\footnotetext[2]{Corresponding author}
\renewcommand{\thefootnote}{\arabic{footnote}}
% \authornote{Corresponding author}
% \authornote{}
\affiliation{%
  \institution{National Engineering Research Center for Multimedia Software, Institute of Artificial Intelligence, School of Computer Science, Wuhan University}
  \city{Wuhan}
  \country{China}}
\email{wangzwhu@whu.edu.cn}
\renewcommand{\shortauthors}{Wan and Wang, et al.}

\begin{abstract}
Large-scale datasets have played a crucial role in the advancement of computer vision. However, they often suffer from problems such as class imbalance, noisy labels, dataset bias, or high resource costs, which can inhibit model performance and reduce trustworthiness. With the advocacy of data-centric research, various data-centric solutions have been proposed to solve the dataset problems mentioned above. They improve the quality of datasets by re-organizing them, which we call dataset refinement. In this survey, we provide a comprehensive and structured overview of recent advances in dataset refinement for problematic computer vision datasets\footnote{All resources are available at \href{https://github.com/Vivian-wzj/DatasetRefinement-CV}{https://github.com/Vivian-wzj/DatasetRefinement-CV}.}. Firstly, we summarize and analyze the various problems encountered in large-scale computer vision datasets. Then, we classify the dataset refinement algorithms into three categories based on the refinement process: data sampling, data subset selection, and active learning. In addition, we organize these dataset refinement methods according to the addressed data problems and provide a systematic comparative description. We point out that these three types of dataset refinement have distinct advantages and disadvantages for dataset problems, which informs the choice of the data-centric method appropriate to a particular research objective. Finally, we summarize the current literature and propose potential future research topics.
\end{abstract}

\begin{CCSXML}
<ccs2012>
   <concept>
       <concept_id>10002944.10011122.10002945</concept_id>
       <concept_desc>General and reference~Surveys and overviews</concept_desc>
       <concept_significance>500</concept_significance>
       </concept>
   <concept>
       <concept_id>10010147.10010178.10010224</concept_id>
       <concept_desc>Computing methodologies~Computer vision</concept_desc>
       <concept_significance>500</concept_significance>
       </concept>
   <concept>
       <concept_id>10002978</concept_id>
       <concept_desc>Security and privacy</concept_desc>
       <concept_significance>100</concept_significance>
       </concept>
 </ccs2012>
\end{CCSXML}

\ccsdesc[500]{General and reference~Surveys and overviews}
\ccsdesc[500]{Computing methodologies~Computer vision}
\ccsdesc[100]{Security and privacy}

\keywords{Dataset refinement, data sampling, subset selection, active learning}

\maketitle

\section{Introduction and Motivation}
\begin{figure}[t]
\begin{center}
\includegraphics[width=0.85\textwidth]{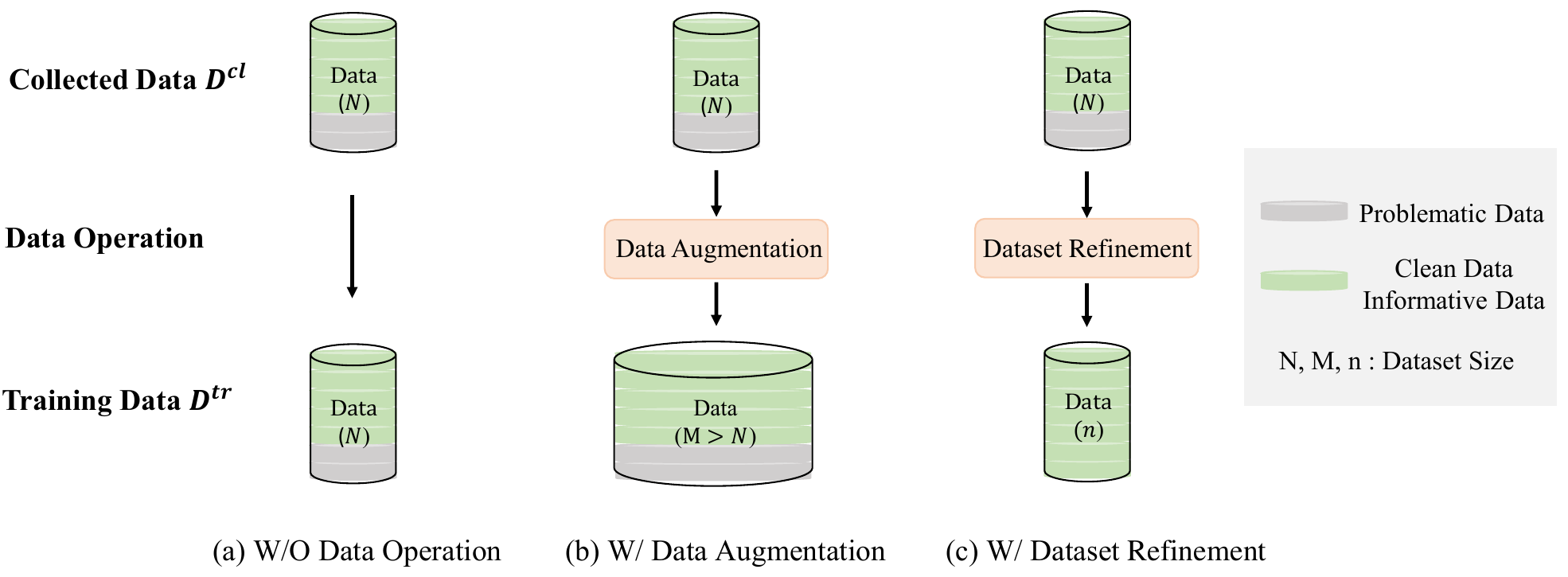}
\end{center}
\caption{Comparison of different data operations. (a) Without (W/O) data operation: typically, once a dataset is collected, it is directly used as a training set without any processing. In this case, the size of the training set $D^{tr}$ is equal to the size $N$ of the collected dataset $D^{cl}$. Problematic data in the collected dataset have received less attention and remain to be addressed. (b) With (W/) data augmentation: in the case of data starvation, data augmentation is often used to make small changes to the collected data as a way to increase the size $M$ and diversity of training data, and it has received increasing attention in recent years. However, instead of being addressed, problematic data increases, inhibiting the effectiveness of augmentation. (c) W/ dataset refinement: it is used to address the problems faced in the collected dataset to improve the dataset quality and, thus, the performance and efficiency of model learning. In most cases, the size of the refined dataset $n \leq N$.}
\label{dataOperations}
\end{figure}

Recently, deep learning has achieved impressive progress in computer vision \cite{liu2021survey, liang2022advances}. The success of deep learning mainly owes to three points, namely advanced deep network architectures (\eg residual networks \cite{he2016deep}), powerful computing devices (\eg Graphic Processing Units (GPUs)), and large datasets with labels (\eg ImageNet \cite{deng2009imagenet}). Among them, deep network architectures and computing devices are well-developed, but obtaining high-quality training datasets is still very difficult. As the attention of researchers was mainly focused on the development and optimization of deep models and computational devices, once the data was ready, it became a fixed asset and received less attention (as shown in Fig.~\ref{dataOperations} (a)). With the development of model architecture, the incremental gains from improving models are diminishing in many tasks~\cite{kiela2021dynabench}. At the same time, relatively minor improvements in data can make Artificial Intelligence (AI) models much more reliable~\cite{liang2022advances}. Therefore, more attention should be paid to data development. Furthermore, there is a prevalent assumption that all data points are equivalently relevant to model parameter updating. In other words, all the training data are presented equally and randomly to the model.
However, numerous works have challenged this assumption and proven that not all samples are created equal \cite{katharopoulos2018not, han2019slimml}. Networks should be aware of the various complexities of the data and spend most of the computation on critical examples. With all these concerns, there is a strong need for the community of artificial intelligence (AI) to move from model-centric research to data-centric research for further improvements in model learning.

Ideally, the collected dataset is correct, without any problem, and can be directly used for model learning. However, since there is no uniform standard for the data collection and labeling process, and the labeling is usually left to crowdsourcing companies rather than labeling experts, the collected dataset is often of low quality, \ie it contains redundant and non-informative data, and often suffers from various problems such as noisy labels \cite{cordeiro2020survey}, class imbalance \cite{kaur2019systematic}, representation bias \cite{li2018resound}, and distribution mismatch \cite{torralba2011unbiased}. For example, ImageNet, the most influential ultra-large benchmark in computer vision, was identified by \cite{beyer2020we} as containing noisy labels. For many computer vision tasks such as face recognition \cite{zhang2017range, cao2020domain}, medical image diagnosis \cite{ju2021relational}, and image classification \cite{van2018inaturalist}, the collected training datasets typically exhibit a long-tailed class distribution, where a minority class has a large number of samples, and other classes have only a small number of samples. Since deep neural networks have the capacity to essentially memorize any characteristics of the data \cite{zhang2021understanding}, those problematic data can drastically inhibit the performance of model training. Besides, the collected data may also face challenges such as compressing the volume and reducing the annotation cost. This is due to the fact that under the advocacy of green AI \cite{schwartz2020green} in the AI community, the training data should not only improve model accuracy but also enable efficient training, thus reducing resource consumption, decreasing AI’s environmental footprint, and increasing its inclusivity. 
Therefore, how to effectively refine the problematic dataset before model training becomes one of the bottleneck problems for a trustworthy and robust AI system.

With the advocacy of data-centric research, some data-centric studies have been conducted for dataset problems in recent years. They aim to make the dataset more accurate and useful by removing irrelevant or redundant data and correcting problems, which we call dataset refinement (as shown in Fig.~\ref{dataOperations} (c)).
Dataset refinement studies can be divided into three main directions according to the refinement process: data sampling, data subset selection, and active learning. 
Data sampling adjusts the frequency and order of training data to promote model training effectively; data subset selection selects the most representative samples from the labeled training set to promote efficient model training; active learning selects the most useful samples from the unlabeled dataset for labeling to minimize the cost of labeling. When comparing them with another type of data operation, namely data augmentation, it can be seen that the latter has received more attention in recent years. This is because data augmentation is a powerful technique to alleviate data-hungry situations in deep learning. Data augmentation \cite{shorten2019survey} increases the amount and diversity of training data by adding small changes to the original data or creating new synthetic data based on the original data, which can improve the performance of models. However, it does not address problematic data, so direct augmentation based on original data can lead to an increase in problematic data (as shown in Fig.~\ref{dataOperations} (b)) and inhibit the effectiveness of augmentation. Therefore, it is recommended to carry out dataset refinement before or after data augmentation and before model training. 

\subsection{Motivation for Work}
The motivation for this comprehensive overview is outlined below:
\begin{itemize}
\item To thoroughly inspect the various problems inherent in or external to computer vision datasets.
\item To focus on the data-centric solution and review the dataset refinement methods for each dataset problem.
\end{itemize}
This article focuses on three main directions in dataset refinement (\ie data sampling, data subset selection, and active learning), and reviews and analyzes their respective developments in light of the problems addressed. This is a data-centric survey that aims to provide insights into dataset refinement methods and to inform researchers and practitioners in the field of computer vision on the selection of appropriate methods to solve specific dataset problems.

\subsection{Our Contributions}
\begin{itemize}
\item The problems faced by computer vision datasets have been elaborated.
\item A comprehensive survey has been conducted to investigate dataset refinement for solving problems in computer vision datasets.
\item Comparative analysis of different dataset refinement methods for solving the same dataset problems has been performed.
\end{itemize}

\begin{table}[t]
\caption{Related Search. They may cover data problems such as Class Imbalance (CI), Noisy Labels (NL), Dataset Biases (DB), High Computational Cost (HCC), and High Labeling Cost (HLC). ``DS'' indicates data sampling; ``SS'' indicates subset selection; ``AL'' indicates active learning. ``-'' means that the corresponding column is not covered.}
\footnotesize
\renewcommand\arraystretch{1.2}
\centering
\begin{tabular}{c|ccccc|cc|ccc}
\hline
\multirow{2}{*}{Research Articles}          & \multicolumn{5}{c|}{Data Problems Covered}    & \multicolumn{2}{c|}{Improvement Perspective}            & \multicolumn{3}{c}{Dataset Refinement}                                                             \\ \cline{2-11} 
                                            & CI & NL & DB & HCC & HLC   & Model-level & Data-level            & DS & SS & AL      \\ \hline
\multicolumn{1}{l|}{\begin{tabular}[c]{@{}l@{}}Data Collection and Quality Challenges in Deep Learning: \\A Data-Centric AI Perspective \cite{whang2023data}  \end{tabular} }     &                                      & \cmark             & \cmark               &                                              &                      &                                  & \cmark & \cmark              & \cmark                &                       \\ \hline
\multicolumn{1}{l|}{\begin{tabular}[c]{@{}l@{}} A Survey on Curriculum Learning \cite{wang2021survey} \end{tabular}}     & \cmark                & \cmark             &                                     & \cmark                       &                       &                                 & \cmark & \cmark              &                                  &                       \\ \hline
\multicolumn{1}{l|}{\begin{tabular}[c]{@{}l@{}}A Review of Instance Selection Methods \cite{olvera2010review} \end{tabular}}   &                                      &                                   &                                     & \cmark                        &                       &                                  & \cmark &                                    & \cmark                 &                       \\ \hline
\multicolumn{1}{l|}{\begin{tabular}[c]{@{}l@{}} DeepCore: A Comprehensive Library for Coreset Selection \\ in Deep Learning  \cite{guo2022deepcore} \end{tabular}  }  &                                      &                                   &                                     & \cmark                        &                       &                                  & \cmark &                                    & \cmark                 &                       \\ \hline
\multicolumn{1}{l|}{\begin{tabular}[c]{@{}l@{}} A Survey on Active Deep Learning: From Model Driven \\ to Data Driven \cite{liu2021survey} \end{tabular}  }      &                                      &                                   &                                     &                                              & \cmark &                                  & \cmark &                                    &                                       & \cmark \\ \hline
\multicolumn{1}{l|}{\begin{tabular}[c]{@{}l@{}} Deep Long-tailed Learning: A Survey \cite{zhang2021deep} \end{tabular}  }       & \cmark                &                                   &                                     &                                              &                       & \cmark            & \cmark & \cmark              &                                       &                       \\ \hline
\multicolumn{1}{l|}{\begin{tabular}[c]{@{}l@{}} Learning from Noisy Labels with Deep Neural Networks: A \\Survey \cite{song2022learning} \end{tabular}  }   &                                      & \cmark             &                                     &                                              &                       & \cmark            & \cmark & \cmark              & \cmark                 &                       \\ \hline
\multicolumn{1}{l|}{\begin{tabular}[c]{@{}l@{}} A Survey on Bias in Visual Datasets \cite{FABBRIZZI2022103552} \end{tabular}  } & \cmark                &                                   & \cmark               &                                              &                       & \multicolumn{2}{c|}{—}      &                 \multicolumn{3}{c}{—}                       \\ \hline
\multicolumn{1}{l|}{Our Survey }                                 & \cmark                & \cmark             & \cmark               & \cmark                        & \cmark &                                  & \cmark & \cmark              & \cmark                 & \cmark \\ \hline
\end{tabular}
\label{tab:related}
\end{table}

\subsection{Related Search}
In this section, we present research relevant to our survey. There is a systematic study~\cite{whang2023data} reviewing the data collection and quality challenges in deep learning from a data-centric AI perspective, which divides the whole machine learning process into data collection, data cleaning, and robust model training. While the survey~\cite{whang2023data} covered the data cleaning associated with dataset refinement, it provided only a cursory summary of data cleaning instead of a detailed summary of how it has developed in recent years. It did not elaborate on the problems in datasets. In addition, there have been some surveys in various directions of dataset refinement (data sampling, data subset selection, or active learning), such as a survey on curriculum learning \cite{wang2021survey}, reviews of instance selection methods \cite{olvera2010review, guo2022deepcore}, and a survey on active deep learning \cite{liu2021survey}. Then, there are surveys~\cite{zhang2021deep, song2022learning, FABBRIZZI2022103552} focusing on one or two data problems but not covering all problems as much as possible. Moreover, they mainly cover methods that overcome data problems at the model level, while less attention is paid to dataset refinement methods. In summary, there is no systematic study that comprehensively reviews and discusses dataset refinement from the perspective of data problems. To fill this gap, we aim to provide a comprehensive survey of recent dataset refinement studies in computer vision conducted before mid-2022. Table~\ref{tab:related} presents the comparison of existing surveys related to problematic data in computer vision.

\subsection{Article Organization}
The survey is organized as follows: we first summarize and describe the various problems inherent in, or external to, computer vision datasets in Section~\ref{sec:summary_problems}. Then, in Section~\ref{sec:viewOptimization}, we provide an overview of three directions of dataset refinement (\ie data sampling, subset selection, and active learning), including their notation, components, definitions, and taxonomy. We classify advanced dataset refinement methods according to the problem solved and the result achieved, and review them in Sections~\ref{sec:DO_RL}, \ref{sec:DO_FL}, \ref{sec:DO_DE}, and \ref{sec:DO_LE}, respectively. We further compare and analyze several related learning problems in Section~\ref{sec:relevant}, including data distillation, feature selection, and semi-supervised learning. Afterward, in Section~\ref{sec:future}, we discuss some of the potential future research directions. Finally, we conclude the survey in Section~\ref{sec:conclusion}.

\section{Summary of Problems in Computer Vision Datasets}
\label{sec:summary_problems}
Generally, due to the lack of uniform standards and specifications for the current data collection and labeling process, published datasets are not perfect~\cite{liang2022advances} and will suffer from some inherent problems, such as class imbalance~\cite{zhang2021deep, Park_2021_ICCV}, noisy labels~\cite{cordeiro2020survey, gui2021towards}, or dataset biases~\cite{FABBRIZZI2022103552, geirhos2018imagenet, torralba2011unbiased}. In addition, with the increasing size of datasets, data collection or application development may face problems such as high labeling costs and excessive computational overhead, which can also inhibit the development of AI. In the following, we will detail the main data problems faced in the field of computer vision and the corresponding research challenges, and briefly describe how to solve them from the perspective of dataset refinement.

\begin{figure}[t]
\begin{center}
\includegraphics[width=0.85\textwidth]{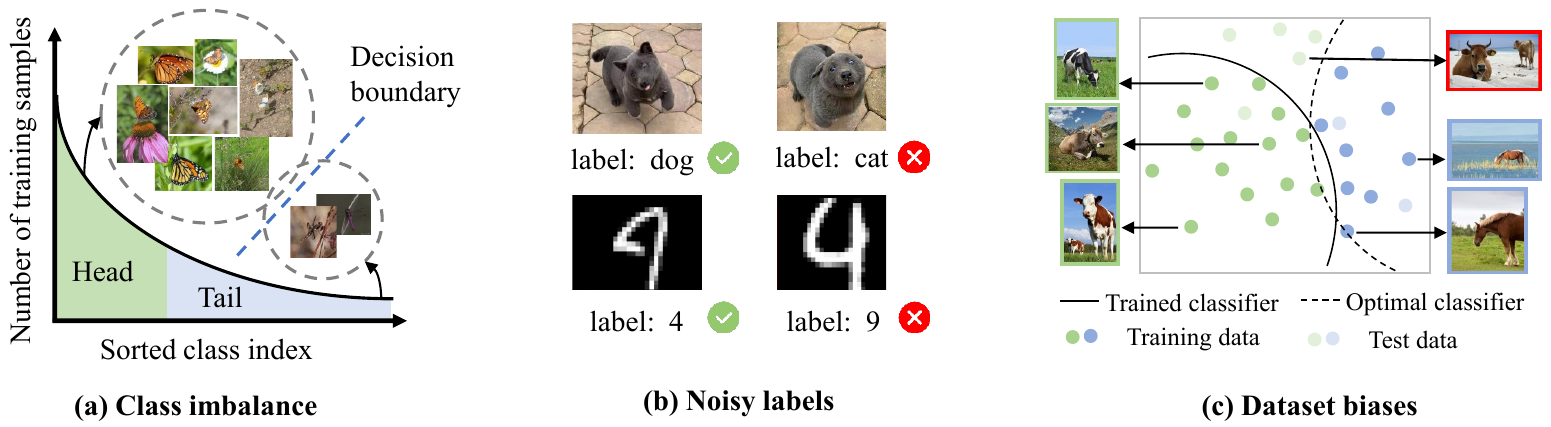}
\end{center}
\caption{Illustrations of the problems inherent in computer vision datasets.}
\label{fig:problems}
\end{figure}

\textbf{Class imbalance}.
Technically, any dataset that shows an uneven distribution between classes can be considered imbalanced \cite{kaur2019systematic, he2009learning}. This issue has attracted considerable interest from academia, industry, and government funding agencies. The main class imbalance of interest to computer vision researchers is the problem of long-tailed class distribution. The long-tailed class distribution (later termed the long-tailed distribution) is a classical class imbalance problem in which a few classes (the head classes) have massive samples while others (the tail classes) have only a small number of samples, as shown in Fig.~\ref{fig:problems}(a). Therefore, in most cases, the class imbalance problem in this article refers to the long-tailed distribution problem. During training, this problem can bias the model toward the head class and cause it to perform poorly on the tail class. As such, it poses a challenge to the robust learning of models. For this problem, data sampling and data subset selection can be used to counteract the long-tail effect by balancing the distribution of head and tail classes and enhancing the learning of tail classes, as summarized in Section~\ref{sec:DS_imbalance} and Section~\ref{sec:SS_imbalance}, respectively.

\textbf{Noisy labels}.
Generally, we default the data to be correctly labeled. However, there may be label issues in the datasets due to inevitable errors by human annotators or automated label extraction tools for images, such as crowdsourcing and web crawling. For example, there are likely at least 100,000 label issues in ImageNet, such as noisy labels that are incorrectly labeled. Some samples with noisy labels from realistic or the MNIST~\cite{lecun1998gradient} dataset are shown in Fig.~\ref{fig:problems}(b). Deep learning with noisy labels is practically challenging, as the capacity of deep models is so high that they can totally memorize these noisy labels sooner or later during model training \cite{han2018co}. As a result, these noisy labels inevitably degenerate the robustness of learned models \cite{nettleton2010study}. For improving the model robustness, there has been a lot of work to overcome this problem by data sampling or data subset selection, as summarized in Section~\ref{sec:DS_Noisy} and Section~\ref{sec:SS_Noisy}, respectively. The noisy labels studied in most of the current work are synthetic random label noise (symmetric or asymmetric label noise \cite{cordeiro2020survey}). In fact, there exists much more instance-dependent label noise~\cite{cheng2020learning} in real-world datasets. For example, annotators could easily label a cat as a lion, but would not easily label a cat as a table. Compared to random labeling noise, instance-dependent labeling noise is more challenging and poorly modeled theoretically and awaits further study.

\textbf{Dataset biases}.
Since most data sets are collected uncontrolled, this leads to dataset biases~\cite{FABBRIZZI2022103552} in the dataset, \ie unintended correlations between inputs and outputs and distribution mismatch between the source and target datasets~\cite{tommasi2017deeper, torralba2011unbiased}. Dataset biases not only inevitably bias the models trained on them but have also been shown to significantly exaggerate model performance, leading to an overestimation of the true capabilities of current AI systems. For example, as shown in Fig.~\ref{fig:problems}(c), in an image classification dataset where ``cow'' always appears in the grass landscape, the model will learn spurious associations between ``cow'' and ``grass landscape''~\cite{beery2018recognition} and perform well on independent and identically distributed (IID) datasets. However, when the model is used to predict images where cows appear outside of typical grassland landscapes (\eg the cow on the beach in the top right corner of Fig.~\ref{fig:problems}(c)), it fails to correctly classify them as ``cow''. This causes the performance of the well-trained model to degrade significantly on adversarial or out-of-distribution (OOD) datasets. To perform fair and trustworthy learning, much work has been done to explore this through data subset selection, as summarized in Section~\ref{sec:DO_FL}.

\textbf{High resource costs}.
With the availability of large amounts of labeled data, deep learning has recently gained significant advances \cite{sun2017revisiting}. However, creating large labeled datasets is often time-consuming and costly. In addition, large-scale training datasets have significantly increased end-to-end training times, computational resource costs, energy requirements \cite{strubell2019energy}, and carbon footprint \cite{schwartz2020green}. To alleviate these problems, extensive work has attempted to reduce labeling costs through active learning or to achieve efficient training through data sampling and data subset selection, as summarized in Section~\ref{sec:DO_DE} and Section~\ref{sec:DO_LE}.

\textbf{Data privacy}.
The current large-scale datasets used for model training are collected directly from the real world, which is expected to simulate real-world data distribution and reduce the bias between training datasets and the real world. However, real data may involve privacy issues, for example, person re-identification datasets contain pedestrian information in public scenes, while vehicle re-identification datasets contain private information such as license plates. With public concern over privacy issues, the collection and use of real datasets have become problematic. It is an urgent and valuable problem to investigate how privacy concerns can be avoided when using real-world data. There is only a small body of work~\cite{kothawade2022prism} that attempts to use data subset selection to protect privacy, awaiting extensive research and application.

\section{Overview of Dataset Refinement}
\label{sec:viewOptimization}

\subsection{Notations}
Fig.~\ref{Fig2} shows an overview of each dataset refinement direction, including data sampling, data subset selection, and active learning. To describe each algorithm procedure more clearly, we explain it in symbols. We first define $D^{cl}$ as the collected dataset. If it is well-annotated, $D^{cl}=\{(x_{i}, y_{i})\}^{N}_{i=1}$, where $x_{i}$ is $i^{th}$ sample and $y_{i}$ is its label (for example, the class label for the image classification task); otherwise, $D^{cl}=\{x_{i}\}^{N}_{i=1}$, where $x_{i}$ is $i^{th}$ sample without the ground truth label. Then we define $D^{tr}$ as the training dataset. Generally, $D^{tr} = D^{cl}$ without any data operation. And we define $D^{vl}$ and $D^{ts}$ as the validation set and test set, respectively, where $D^{vl}=\{(x_{k}, y_{k})\}^{K}_{k=1}$ and $D^{ts}=\{(x_{j}, y_{j})\}^{J}_{j=1}$. For computer vision tasks, its goal is to search a mapping function $f_{\theta}: X \rightarrow Y$, where X is the input space, Y is the output space, and $\theta$ is the set of parameters that are trained by $D^{tr}$. The function $f_{\theta}$ is an element of the space of possible functions, usually called the hypothesis space. The loss function $l(f_{\theta}(x_{i}), y_{i})$ is usually used to evaluate the performance of the model on data $(x_{i}, y_{i})$. Higher loss values indicate poor performance of the model in predicting the label of the data $(x_{i}, y_{i})$. The empirical risk on the full training dataset with labels is expressed as:
\begin{equation}
R(f_{\theta})=\frac{1}{N}\sum^{N}_{i=1}l(f_{\theta}(x_{i}),{y_{i}}).
\label{ER_function}
\end{equation}

Existing model optimization methods facilitate model learning by minimizing the empirical risk subjected to network parameters:
\begin{equation}
\hat{\theta} =\arg\min_{\theta}R(f_{\theta}).
\label{update_function}
\end{equation}

If problematic data exist in the collected dataset, there is no doubt that the model parameters will be optimized in a biased direction~\cite{algan2021image}, which can lead to dramatic performance degradation of the well-trained model in real application scenarios~\cite{geirhos2020shortcut}, a terrible direction for critical domains like healthcare, transport, and finance. To address this, a large number of dataset refinement methods have been proposed and will be discussed further in the subsequent sections. After learning from $D^{tr}$, the function $f_{\theta}$ will be tested on the $D^{ts}$ to check its effectiveness.
%geirhos2020shortcut

\subsection{Components}
\begin{figure}[t]
\begin{center}
\includegraphics[width=0.65\textwidth]{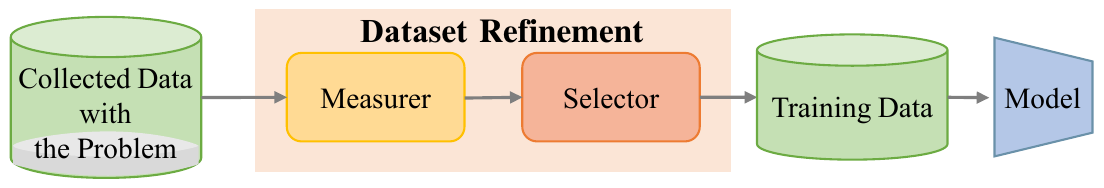}
\end{center}
\caption{Dataset refinement process.}
\label{DO}
\end{figure}

Dataset refinement improves computational efficiency and training effectiveness from the data level and breaks AI research bottlenecks by shifting the focus from big data to good data. As shown in Fig.~\ref{DO}, the process of dataset refinement consists of two main components, namely the measurer and the selector. We define $M_{\gamma}$ and $S_{\delta}$ as the measurer and selector, respectively.

The \textbf{Measurer} is used to measure the score of each data, which can indicate the importance or difficulty degree of data. The key to this is the metrics. Existing works have empirically explored different metrics according to the status of the prediction results, such as influence function-based metrics, representation-based metrics, confidence-based metrics, uncertainty-based metrics, and loss-based metrics, which are commonly used in dataset refinement. (1) \textit{Influence function}~\cite{koh2017understanding, wang2020less}, a data valuation metric, aims to approximate the change of test risk on a given test distribution $Q$. It uses the gradient of the loss function with small perturbations $\epsilon$ to estimate the data value, \ie the importance of the data for the task at hand. It is a particularly general approach as many models can be cast in the framework, and it can work with non-differentiable objectives. (2) \textit{Representation-based metrics} \cite{farahani2009facility} measure the value of each data in the feature space. Generally, it is assumed that data points close to each other in the feature space tend to have similar properties, resulting in redundant information. (3) \textit{Confidence}, the predicted probability of the classifier on the true class, can be used to measure the impact of per data on model optimization. Some work on robust learning \cite{northcutt2017learning, northcutt2021confident} empirically suggests that samples with lower confidence may contribute more to model learning than samples with higher confidence. By recording and analyzing the changes in the confidence of per data during the training, (4) \textit{Uncertainty} \cite{settles2012active, coleman2019selection, sun2020crssc} is used to measure the consistency of label prediction. (5) \textit{Loss} or \textit{gradient} of each sample is also usually used to measure the importance. If a sample contributes more to errors or losses when training a neural network, then it is more important. For some metrics, the formulas for calculating their measurement scores are shown in Table~\ref{tab:formula}. In addition to the above hand-designed metrics for measures, there are learning-based measures that utilize meta-learning techniques or policy networks to re-weight or select training data automatically. Different metrics are designed and selected depending on the specific problem.

\begin{table}[t]
\centering
\caption{The formula of some metrics.}
\begin{tabular}{l|c}
\hline
Metrics                  & Formulas\\ \hline
Influence function \cite{koh2017understanding}& $\begin{aligned} \phi\left(x_{i} \sim P, x_{j} \sim Q\right) &\left.\triangleq \frac{d l_{j}\left(\hat{\theta}_{\epsilon}\right)}{d \epsilon}\right|_{\epsilon=0} \\ &=-\nabla_{\theta} l_{j}(\hat{\theta})^{\top} H_{\hat{\theta}}^{-1} \nabla_{\theta} l_{i}(\hat{\theta}) \end{aligned}$ \\& where 
% $\hat{\theta}_{\epsilon} \triangleq \arg \min _{\theta} \frac{1}{n} \sum_{i=1}^{n} l_{i}(\theta)+\epsilon l_{j}(\theta)$ and 
$ H_{\hat{\theta}} \triangleq \frac{1}{n} \sum_{i=1}^{n} \nabla_{\theta}^{2} l_{i}(\hat{\theta})$ \\  \hline
Confidence         & $\phi_{confidence}(x) = P(y=y_{true}| x, \theta)$         \\ \hline
Uncertainty based -1 \cite{ruzicka2020deep}      & $\phi_\textit{least confidence}({x})=1-\max _{i=1, \ldots, C} P({y}=i| {x}, \theta)$         \\ \hline
Uncertainty based -2  \cite{joshi2009multi}     &  $\phi_{\textit{margin}}(x)=1-\min(P(y_{1}|{x}, \theta)-P(y_{2}|{x}, \theta))$        \\ \hline
Uncertainty based -3 \cite{settles2012active}      & $\phi_{\textit {entropy}}({x})=-\sum_{i=1}^{C} P({y}=i | {x}, \theta) \log P({y}=i | {x}, \theta)$         \\ \hline
Uncertainty based -4 \cite{sun2020crssc}    &   $\phi_{\textit {variance }}({x})=\sqrt{\frac{1}{C} \sum_{i=1}^{C}(P(y=i|x,\theta)-\mu)^{2} }$      \\ \hline
\end{tabular}
\label{tab:formula}
\end{table}

The calculated measurement scores are used to re-weight per data during training. At this point, the empirical risk on the full training set with labels can be expressed as:

\begin{equation}
R(f)=\frac{1}{N}\sum^{N}_{i=1}(W(x_{i}, y_{i})l(f_{\theta}(x_{i}), {y_{i}})).
\end{equation}
The weight value $W$ is different for each data. Useful data are assigned high weights, while problematic data are assigned low weights. For example, when training on a dataset with noisy labels, ideally, the weight of the noisy labeled data is 0, and the weight of the clean data is 1. In addition, the range of weights taken for each dataset refinement direction is different. The weights in data sampling are variable, while in data subset selection and active learning, the weights are taken in the binary range of 0 or 1. Here, 0 indicates that the data is removed during training, and 1 indicates that the data is selected for training.

The \textbf{Selector} is the strategy that decides when and how to select the data according to the score. It can be used before the final training of the target model to refine the overall structure of the training data or in each training iteration to optimize the selection of the batch data.

\begin{figure}[t]
\begin{center}
\includegraphics[width=0.85\textwidth]{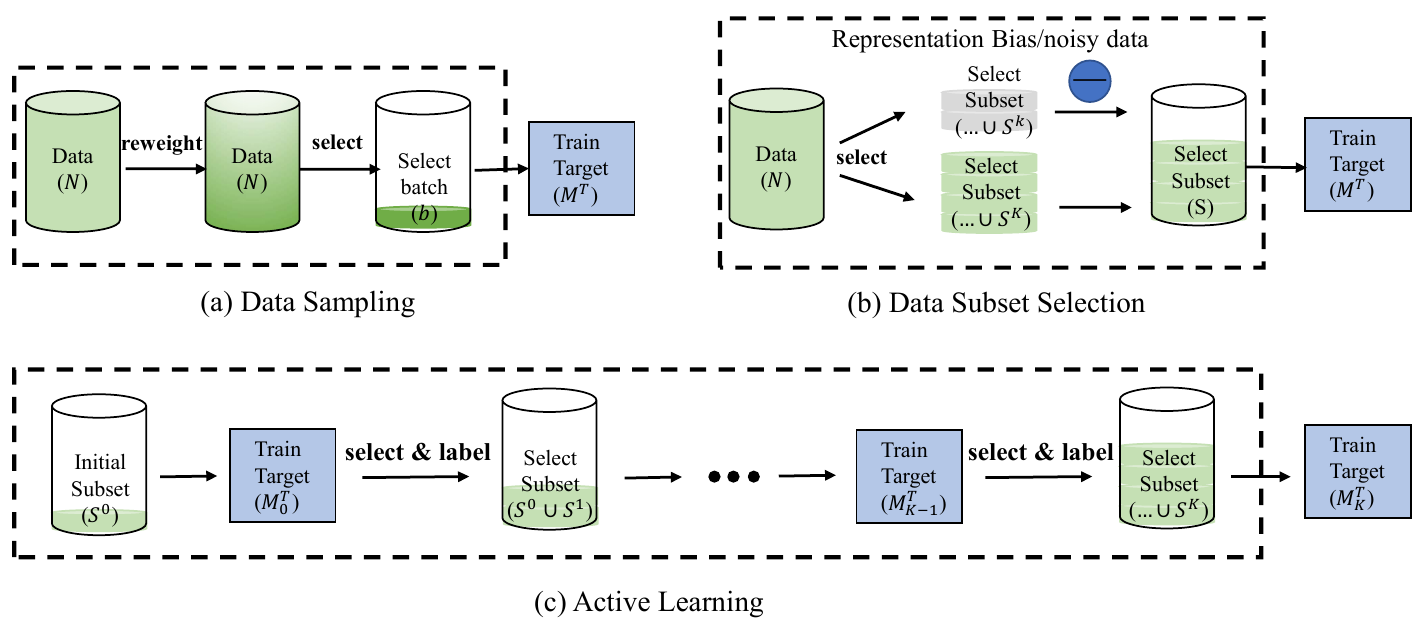}
\end{center}
\caption{Overview of each dataset refinement direction.}
\label{Fig2}
\end{figure}

It is the key point of dataset refinement to effectively design the measure and selector. Depending on the state (quantity, the time point of being selected, \etc.) of the data to be selected, dataset refinement studies can be divided into three directions: data sampling, data subset selection, and active learning. Their data flows are shown in Fig.~\ref{Fig2}. They have different algorithm design structures and are applicable to solve different data problems. In the following, each direction is described in detail.

\subsection{Dataset Refinement Directions}
\label{data_optimizations}

\textbf{Data sampling}.
It investigates how to adjust the frequency and order of data sampled from the whole collected data source \cite{sun2021autosampling}, which aims to promote model convergence, achieve better model performance \cite{zhang2021one}, or balance the dataset so that a fair model can be trained \cite{li2019repair, chawla2002smote}. The dataset refinement procedure contains two core stages, namely the re-weighting stage and the selection stage. Initially, the training data $D^{tr}=D^{cl} \in P$, where $P$ refers to the training distribution. Then the training distribution $P$ is re-weighted to $W*P$, where the weights $W$ are decided by the score calculated by $M_{\gamma}$. After that, the selector decides the sequence of data subsets throughout the training process based on the re-weighted distribution. At each iteration, the most suitable data are selected to form a batch $b$ for training. 

\textbf{Data subset selection (subset selection)}.
It selects the most representative samples from the labeled dataset for training, aiming at minimizing training time and computational resources and increasing efficiency while maintaining performance \cite{killamsetty2021retrieve, han2019slimml} or accelerating hyper-parameter tuning \cite{killamsetty2022automata}. It benefits not only data-efficient learning but also other downstream tasks such as active learning, continuous learning, and neural structure search. There are two ways to select data. In the first one, the selector $S_{\delta}$ is designed to directly select representative samples that constitute the final training dataset $D^{tr}$. In the second one, the selector $S_{\delta}$ is used as a filter to filter out the biased or noisy subset $S^f$, and the remaining data is used as the final training dataset $D^{tr}$, where $D^{tr} = D^{cl} \setminus S^f$ with the constraint $|D^{tr}| < |D^{cl}|$. After being trained on the selected data $D^{tr}$, the model $f(\tilde{\theta})$ aims to achieve a performance close to or better than that of the model $f(\hat{\theta})$ trained on the original full dataset $D^{cl}$.

\textbf{Active learning (AL)}.
It aims to select the most useful samples from the unlabeled dataset for labeling, thus minimizing the cost of labeling while maintaining performance. In the initial learning, the training data $D^{tr}$ is a subset sampled from the collected dataset $D^{cl}=\{x_{i}\}^{N}_{i=1}$ and labeled, then the remaining unlabeled samples are used as the candidate dataset $D^{cd}$, \ie $D^{cl} = D^{tr} \cup D^{cd}$. And $|D^{tr}| \ll |D^{cl}|$. In subsequent an arbitrary \textit{k}-th training iteration, the measurer is designed to measure the validity of each unlabeled data in $D^{cd}_{k}$, and then the selector $S_{\delta}$ is used to select some of the most valid samples $S^k$. After selection, $S^k$ will be labeled, and then the function $f_{\theta}$ is re-trained on the new $D^{tr}_{k}$ where $D^{tr}_{k} = D^{tr}_{k-1} \cup S^k$ and $D^{cd}_{k+1} = D^{cd}_{k} \setminus S^k$.

\begin{table}[b]
\caption{Main notation.}
\begin{tabular}{c|l}
\hline
$D^{cl}$, $D^{tr}$, $D^{ts}$, $D^{vl}$  & collected dataset, training set, testing set and validation set \\
$x_{i}$, $x_{j}$, $x_{k}$  & samples for training, testing and validation respectively, $\in \mathbb{R}^{d}$ \\
$y_{i}$, $y_{j}$, $y_{k}$  & labels of training, testing and validation sample \\
$N, M, n, J, K$& the size of dataset $D^{cl}$, $D^{tr}$ (W/ data augmentation), $D^{tr}$ (W/ dataset refinement), $D^{ts}$, and $D^{vl}$\\
$\theta$ $(\hat{\theta}, \tilde{\theta})$  & learned network parameters (trained on $D^{cl}$, and trained on the refined training set), $\in \mathbb{R}^{d}$ \\
$l_{i}\left(\theta\right)$, $l_{j}\left(\theta\right)$  & model $\theta$'s loss on the training sample $x_{i}$ and the testing sample $x_{j}$, $\in \mathbb{R}$\\
$W, W(x_{i}, y_{i})$ & weight, the weight value of data $(x_{i}, y_{i})$, $\in \mathbb{R}$\\
$P(y=i|x, \theta)$ & predicted probability
of sample $x$ belonging to the $i$-th class, $\in \mathbb{R}$\\
$P$  & training distribution \\
$Q$  & a specific test distribution\\ 
$\epsilon$ & small perturbations used in \textit{Influence function}\\
$H_{\hat{\theta}}$ & the Hessian matrix based on full set risk minimizer\\
$C$ & the number of classes, $\in \mathbb{R}$ \\
$\mu$ & the mean value of predicted probabilities, $\in \mathbb{R}^{+}$\\
$M_{\gamma}$  & measurer \\
$S_{\delta}$  & selector \\
\hline
\end{tabular}
\label{notation}
\end{table}

\subsection{Taxonomy}
\begin{figure}[!t]
\begin{center}
\includegraphics[width=1\textwidth]{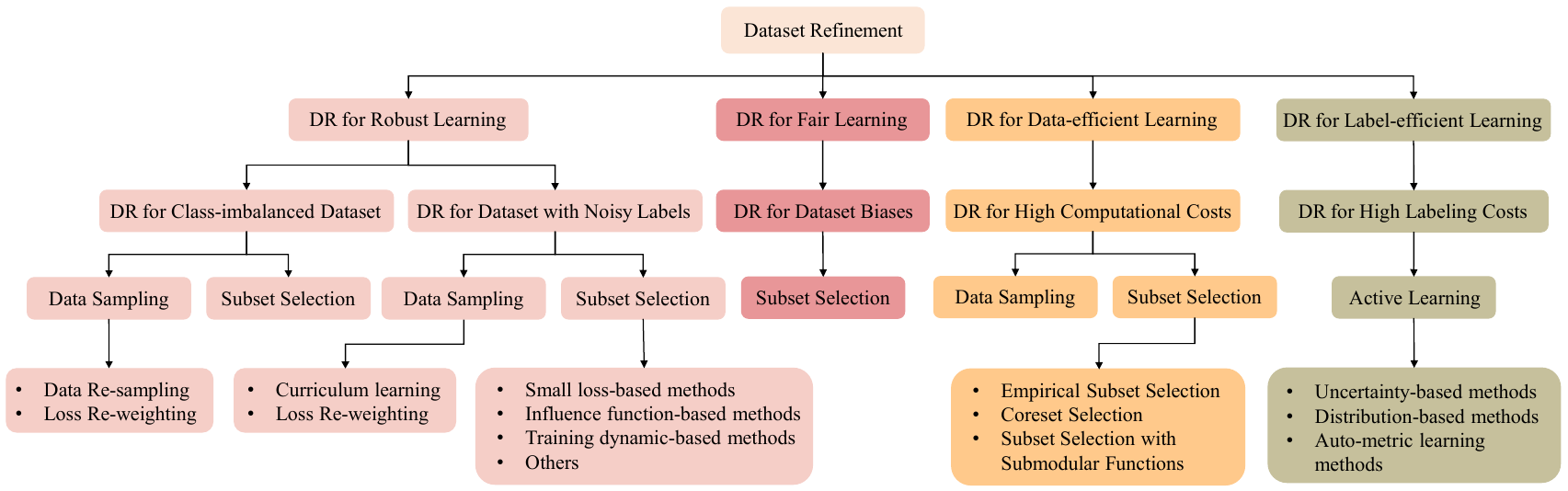}
\end{center}
\caption{Problem based taxonomy of dataset refinement (DR) methods.}
\label{fig:Taxonomy}
\end{figure}

According to Section~\ref{sec:summary_problems}, it can be known that there are some problems in the dataset that can hinder the training of the model. To solve these data problems, the computer vision community has explored dataset refinement. As shown in Section~\ref{data_optimizations}, we classify dataset refinement methods into three directions, namely data sampling, data subset selection, and active learning. Each direction can be used to solve different data problems, \eg data sampling can be used to solve class imbalance, dataset biases, and noisy labels. In addition, it can achieve different effects by solving different data problems, \ie solving the inherent problems of data, such as class imbalance and noise labeling, can improve the robustness of the model. According to the taxonomy of this article shown in Fig.~\ref{fig:Taxonomy}, we classify the problems according to the achieved effects, and later review the dataset refinement methods in detail according to each problem.

\section{Dataset Refinement for Robust Learning}
\label{sec:DO_RL}
Robust Learning on problematic datasets has been well studied \cite{northcutt2017learning, ren2018learning, song2019selfie, sun2020crssc, song2021robust}. Here, we only review prior dataset refinement works that deal with data problems. For a work on dealing with noisy labels based on model optimization, one can refer to the survey~\cite{song2022learning}. For the work on overcoming the class imbalance based on model optimization, one can refer to the survey~\cite{zhang2021deep}.

\subsection{On the Class-imbalanced Dataset}
\label{dataset_CI}
\subsubsection{Data Sampling}
\label{sec:DS_imbalance}
Data sampling has been one of the most widely used methods to address the problem of class imbalance over the past few decades \cite{chawla2002smote, wang2019dynamic}. Other non-dataset-refinement methods for addressing the class imbalance problem can refer to surveys \cite{kaur2019systematic, zhang2021deep}.

\textbf{Data re-sampling.} The basic and simple methods of data sampling are random under-sampling and random over-sampling \cite{drummond2003c4}. Random under-sampling balances the class distribution by randomly excluding data from the head classes. In terms of data flow, this sampling method also belongs to data subset selection. Random over-sampling balances the class distribution by randomly and repeatedly selecting samples from the tail classes to increase the sample size of the tail classes. However, when faced with the extreme class imbalance (\ie the long-tailed distribution), random over-sampling can lead to overfitting of the model to the tail classes and degrade the performance of the model on the head classes. Recently, various sampling methods have been proposed in computer vision tasks to overcome the long-tailed distribution problem, and they use different information to weigh and sample the samples. Bi-level class balanced sampling approach \cite{wang2020devil} was designed to solve the long-tailed instance detection and segmentation problems. In particular, in addition to image-level sampling, an instance-level sampling strategy was designed to alleviate the class imbalance in instance segmentation. Faced with long-tailed video recognition, \citet{zhang2021videolt} proposed the FrameStack method that performed frame-level sampling to balance class distributions. Moreover, the sampling rates for each class are dynamically adjusted based on the running model performance during training. Additionally, there are some methods that bootstrap data sampling with the help of an additional balanced dataset \cite{NEURIPS2020_2ba61cc3, Zang_2021_ICCV}. For example, \citet{NEURIPS2020_2ba61cc3} proposed a meta-learning-based sampling method that learned the best sample distribution parameter by optimizing the model classification performance on a balanced meta-validation set. Similarly, \citet{Zang_2021_ICCV} proposed to utilize the classification loss on a balanced meta-validation set to adjust the feature sampling rate for different classes. 

\textbf{Loss re-weighting.}
There have also been studies that soft-sample the data by re-weighting the sample losses, generally called cost-sensitive learning. To balance the class distribution from the loss level, it generally increases the loss weights of the tail classes and decreases the loss weights of the head classes, thus alleviating the model optimization bias towards the head classes. Therefore, the key point of loss re-weighting methods is to investigate how to measure the loss weights of each class. Most directly, we can calculate the loss weights using the class label frequencies of the training samples, the influence of the samples \cite{Park_2021_ICCV}, or the distribution consistency between the model predictions and the balanced reference class distribution \cite{Zhang_2021_CVPR}. Taking the class label frequency $fr$ of the training samples as an example, the loss weight for class $i$ is the inverse class frequency, \ie $W_{i} = \frac{1}{fr_{i}}$. Thus, a higher $fr$ corresponds to a lower loss weight. In addition, observing that class imbalance usually increases the prediction difficulty of tail classes, whose prediction probabilities are lower than those of head classes, \citet{lin2017focal} designed focal loss, a classical method for solving class imbalance, which uses prediction probabilities to inversely re-weight the loss for each class. Since reducing the weights of majority classes would cause the hard samples in majority classes to become harder to learn and hurt the overall performance, \citet{yu2022re} proposed an instance-level re-balancing strategy to dynamically adjust the sampling probabilities of instances according to their difficulty. Apart from the hand-designed weighting algorithms described above, there are also methods to investigate how to learn the class weighting function from the data automatically. Meta-learning is a powerful way to do this. Based on it, \citet{shu2019meta} proposed the Meta-Weight-Net, a parameterized weighting function whose training is guided by a small balanced validation set. Finally, Meta-Weight-Net can fit a proper weighting function well in the class-imbalanced dataset. 

\subsubsection{Subset Selection}
\label{sec:SS_imbalance}
For dealing with extremely imbalanced cases, \ie long-tailed distribution datasets, it is intuitive to balance the number of training samples of different classes during model training. An easy and common subset selection method is the random under-sampling \cite{ng2002combining}, which randomly selects and discards the samples from head classes and discards them. Similarly, it was observed that the effects of input data points on model parameter updates show a long-tailed distribution, \ie most data points (head data) have much smaller effect values than a small fraction of data points (tail data), so \citet{han2019slimml} proposed to remove non-critical data with effect values smaller than a preset threshold. However, subset selection methods tend to discard a large amount of data, which means that a lot of valid learning information is lost and would degrade the performance of the model in head classes. Therefore, few subset selection methods have been designed to solve the problem of long-tailed distribution.

\subsection{On the Dataset with Noisy Labels}
\label{Dataset_NL}
\subsubsection{Data Sampling}
\label{sec:DS_Noisy}
Data sampling is also one of the widely used methods to deal with noisy labels. It assigns different weights $W$ to the training samples, which aims to emphasize cleaner data and mitigate the negative impact of harmful data on the generalization ability of the model for better updating of the model parameters. 

\textbf{Curriculum learning.} Inspired by the meaningful learning order in the human curriculum, curriculum learning (CL) \cite{bengio2009curriculum}, a classical data sampling strategy, proposes a meaningful sequence of data to train a model, \ie smaller and easier data to larger and harder data to train a machine learning model. In the CL setting, the data with noisy labels correspond to harder examples in the datasets, while the cleaner data form the easier part. The CL strategy encourages training on easier data, so the model wastes less time on more difficult and noisy examples to achieve effective learning. Several methods are proposed to tackle the problem of noisy labels by exploiting the merits of CL. Generally, it is not known a priori which samples are easy, so the key point of CL is to distinguish between easy data and hard/noisy data. The most intuitive way is to use the confidence of the label predicted by the model, \ie the label prediction probability, to perform the distinction. Higher confidence means easier data. In self-paced learning \cite{kumar2010self}, it relies on training loss to identify noisy data from the full training set. In the empirical studies \cite{arpit2017closer}, it has been observed that Deep Neural Networks (DNNs) tend first to learn simple and generalized patterns and then gradually overfit to all noisy patterns, which is called the \textit{memorization effect}. Thus, noisy data tends to have higher losses in early training \cite{song2019does}. Thus, examples with smaller losses on the current model are considered to be easier and can be used in early training. Since it focuses more on easy samples, CL is thought to slow down learning, and \cite{chang2017active} suggests choosing uncertain samples with high prediction variances. Based on the CL strategy, CurriculumNet \cite{guo2018curriculumnet} was proposed for training a model from massive datasets with noisy labels. This method measures the noisy level of data using its distribution density in a feature space. It first clusters the training data in the feature space and identifies samples belonging to low-density clusters as those more likely to be mislabeled. The data are then presented to the model for training in a meaningful order in the sequence. 

\textbf{Loss re-weighting.} 
Apart from the hard-sampling way that removes noisy data ($W = 0$) and selects other data ($W = 1$) in a meaningful order for training, the noisy label problem can also be overcome by the soft-sampling way, \ie by adjusting the weight of each data loss value to optimize the model training. Motivated by the importance re-weighting method \cite{liu2015classification}, loss re-weighting aims to assign smaller weights to the examples with noisy labels and greater weights to clean examples. The key point of the loss re-weighting method is to study how to weigh per data. Active bias \cite{chang2017active} focused on uncertain examples with inconsistently labeled predictions and assigned their prediction variances as loss weights for training. \citet{wang2017multiclass} proposed to use the ratio of two joint data distributions $W(x, y) = P_{clean}(x, y) / P_{noisy}(x_{i}, y_{i})$ to determine the loss weight of each noisy data. \citet{zhang2021dualgraph} proposed to re-weight the examples according to the structural relations among labels, which could eliminate the abnormal noise examples. Besides, learning-based re-weighting methods also gain more and more attention for label noise-robust learning. \citet{ren2018learning} proposed a novel meta-learning algorithm that performed a meta gradient descent step to minimize the loss on a clean and unbiased validation set, which is believed to learn the best sample weighting.

\subsubsection{Subset Selection}
\label{sec:SS_Noisy}
Subset selection methods aim to filter out noisy labeled data and ensure that only clean data (\ie samples with uncorrupted labels) are used for training. The key issue in subset selection is how to identify noisy samples \cite{brodley1999identifying}. There are many explorations in noisy label identification, which can be mainly divided into small loss-based, influence-based, training dynamic-based, and others.

\textbf{Small loss-based methods.} According to the \textit{memorization effect} \cite{arpit2017closer} and observation of work \cite{song2019does}, plenty of works~\cite{jiang2018mentornet, shen2019learning, yang2022mutual} assume that samples with small losses are clean and identify clean samples out of the mini-batch based on their forward losses, which is called the small-loss trick. For example, MentorNet \cite{jiang2018mentornet}, a loss-based subset selection method, pre-trains an additional teacher network to supervise the training of a student network. During training, the teacher network selects data with small losses, which are likely to be correctly labeled, and then provides these clean data to the student network. Co-teaching \cite{han2018co}, a new deep learning paradigm, combats noisy labels by using two peer networks. Each network selects its small-loss samples as clean samples and then feeds these clean samples to its peer network for further training. Co-teaching+ \cite{yu2019does} further employs the disagreement strategy of \cite{malach2017decoupling} based on Co-teaching. INCV \cite{chen2019understanding} randomly divides noisy training data and employs cross-validation to identify the clean data. After that, the Co-teaching strategy is adopted to train the DNN on the identified examples. ITLM \cite{shen2019learning} iteratively minimizes the trimmed loss by alternating between selecting a fraction of small-loss samples at the current moment and retraining the DNN using them. \cite{lyu2019curriculum} also relies on the ``small loss'' heuristic, but the threshold for sample selection is adapted based on the knowledge of label noise rates. However, the subset selection process in these methods is generally based on heuristics without guarantees. For this, \citet{gui2021towards} formalized the small-loss criterion to better tackle noisy labels and theoretically explained why this trick works.

However, the selected clean samples tend to be easy, resulting in numerous hard samples that contribute to accurate and robust model training being ignored. To further filter out the useful hard samples, SELFIE \cite{song2019selfie} introduces the ``uncertainty'' for selection, where the samples with low uncertainty are regarded as the hard samples. Then, they correct the losses of the hard samples and combine them with the losses of clean samples to propagate backward. MORPH \cite{song2021robust}, a novel self-transitional learning scheme, estimates the optimal transition point in the seeding phase and iteratively expands an initial safe set into a maximal safe set via self-transitional learning in the evolution phase.

Small loss-based methods assume that many correctly labeled examples tend to exhibit smaller losses than incorrectly labeled examples, \ie in terms of distribution, correctly labeled examples and incorrectly labeled examples have less overlap in the loss distribution (symmetric label noise). Therefore, these methods will fail when the loss distributions of the correctly labeled examples and the incorrectly labeled examples overlap mostly (asymmetric)~\cite{song2021robust} and when there exists instance-dependent label noise.

\begin{figure}[ht]
\begin{center}
\includegraphics[width=0.5\textwidth]{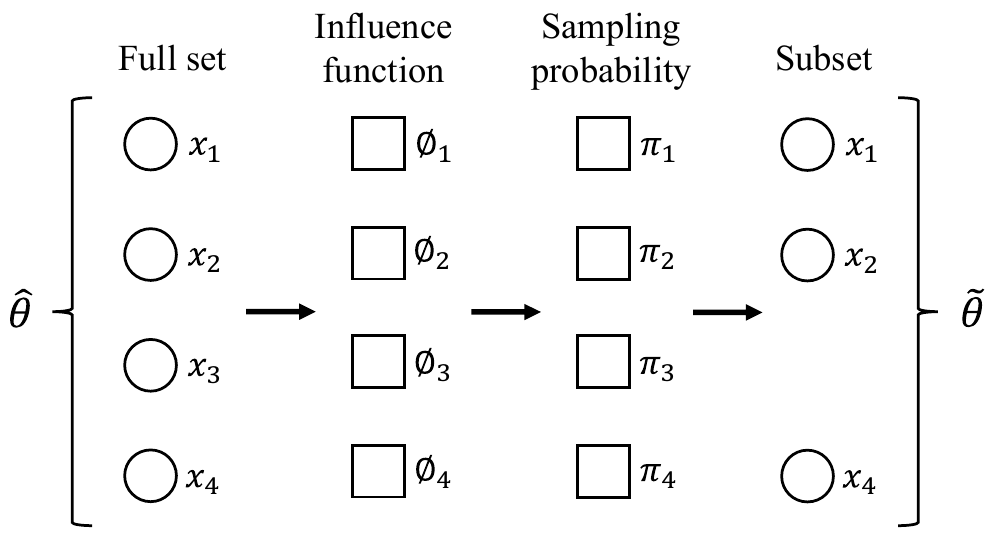}
\end{center}
\caption{Influence function based subset selection framework \cite{wang2020less}.}
\label{IF}
\end{figure}

\textbf{Influence function (IF)-based methods.} The small-loss trick typically holds for deep models but not for any predictive models \cite{zhang2021understanding}, while influence-based subset selection methods can overcome this limitation. Using the influence function \cite{cook1980characterizations}, the methods can estimate the influence of each training sample on the model's predictions in the test set. Any training sample that causes an increase in the test loss is considered harmful and will be removed later. The selection framework is shown in Fig~\ref{IF}. It shows four steps and contains two rounds of training. $(1)$ The first round of training: train the model on the full training set and obtain $f_{\hat{\theta}}$. Here, $\hat{\theta}$ denotes the learned network parameters, and $\hat{\theta}=\arg \min _{\theta} \frac{1}{N} \sum_{i} l\left(x_{i}, \theta\right)$. $(2)$ Compute IF according to the Table~\ref{tab:formula}. $\vec{\phi}=\left(\phi\left(x_{1}, \hat{\theta}\right), \phi\left(x_{2}, \hat{\theta}\right), \ldots, \phi\left(x_{N}, \hat{\theta}\right)\right)$. $(3)$ Compute sampling probability. $\vec{\pi}=\left(\pi\left(\phi_{1}\right), \pi\left(\phi_{2}\right), \ldots, \pi\left(\phi_{N}\right)\right)$. $(4)$ The second round of training: train the model on the selected subset and obtain $f_{\tilde{\theta}}$. Similarly, $\tilde{\theta}$ represents the learned network parameters, and $\tilde{\theta}=\arg \min _{\theta} \frac{1}{\left|\left\{i, o_{i}=1\right\}\right|} \sum_{o_{i}=1} l\left(x_{i}, \theta\right)$.

Following this framework, \citet{wang2020less} proposed an unweighted subsampling to draw a subset from the original dataset and potentially achieve better performance on the test dataset. \citet{wang2018data} computed the influence of removing each training example on the loss across all validation samples and improved the model's accuracy by dropping unfavorable training samples, whose influence value is positive. Influence-based methods are agnostic to a specific model but require the computation of Hessian values for the DNN, which is very costly.

\textbf{Training dynamic-based methods.} Instead of selecting clean data based on the metrics observed during training, the training dynamics-based methods are based on metrics that describe how the data changes over the training times or iteration rounds. These metrics are continuously recorded during training and then calculated at the end of training for subsequent selection of clean data. Various options have been proposed, such as forgetting events and confidence learning.

During the model training, if a sample is correctly classified by the model and then incorrectly classified again as the model parameters are updated, it is said that a \textit{forgetting event} has occurred for that sample. According to work \cite{toneva2018empirical}, it is known that samples with noisy labels tend to experience more forgetting events than normal samples during the model training. Based on this heuristic rule, we can record the total number of forgetting events experienced by each sample and further identify the possible noisy labeled data. Inspired by the work, MORPH \cite{song2021robust} robustly updates the DNN with the current maximal safe set and then derives a more refined set by adding newly memorized samples and deleting newly forgotten samples from the maximal safe set.

\textit{Confidence learning} is an uncertainty-based method. Uncertainty is the variance in the predicted probability of the correct class across iterations, and is used to measure the consistency of label prediction. Rank Pruning \cite{northcutt2017learning}, a confidence learning-based method designed for binary classification, first estimates noise rates using predicted probability ratios with the help of a probabilistic classifier, and then relies on the output confidence of the base network in the training data to remove the most unconfident samples based on the estimated noise rate. Here, confidence is the prediction probability. High confidence means that the prediction probability is close to 1 when the label is positive and 0 when the label is negative. For the multiclass classification tasks, \textit{confidence learning} \cite{northcutt2021confident} is further proposed, which is based on the assumption of the classification noise process that noisy labels are class-conditional and depend only on the true class, not the data. Data with noisy labels are identified by estimating the conditional probability between a given noisy label and the potentially correct label.

Based on the idea of small loss, O2U-Net \cite{huang2019o2u} records the loss of every sample during the cyclical training (from overfitting to underfitting). Then, clean data with small averaged normalized losses are selected for model re-training from scratch, while false-labeled examples with large losses are removed. Observing that clean samples have a larger margin than noisy samples, \citet{pleiss2020identifying} introduced the AUM metric to separate correctly-labeled data from mislabeled data, where AUM measures the average difference between the logit values for a sample’s assigned class and its highest non-assigned class. 

Since the calculation is performed after the target model training, the metrics used in the training dynamics-based methods are stable. Nevertheless, the computational cost of the subset selection process inevitably increases linearly with the number of training rounds.

\textbf{Others.} 
Consistency of networks can also be used to identify the mislabeled data \cite{malach2017decoupling, nguyen2019self, karim2022unicon}. For example, \citet{malach2017decoupling} maintained two networks, and they assumed that the label was noisy if both networks disagreed with the given label. And the networks are updated using the samples with different label predictions. \citet{karim2022unicon} used the Jensen-Shannon divergence (JSD) to measure the disagreement between the two networks and estimated the filter rate from the JSD distribution. Based on the filter rate, they filtered out an equal number of noisy samples from each class for a uniform selection of easy and hard samples. Apart from relying on the prediction of models, subset selection can also be performed by analyzing data representations. TopoFilter \cite{wu2020topological} utilizes the spatial topological pattern of learned representations to detect true-labeled examples with theoretical proof. NGC \cite{wu2021ngc} iteratively constructs the nearest neighbor graph using latent representations and performs geometry-based sample selection by aggregating neighborhood information. Soft pseudo-labels are assigned to unselected examples. Besides, noisy data can also be identified by specially designed networks. For example, \citet{nagalapatti2022your} designed the Relevant Data Selector (RDS) module, which is private to each client in the federated learning setting. The RDS module assigns low values to irrelevant/noisy data to identify and filter them and improve the performance of the global learning model.

For the application of fine-grained vision learning, web images have received increasing attention because they are easily accessible and do not require manual annotation \cite{yao2016domain}. However, due to the presence of noisy labels, models trained directly on web images tend to have an inferior recognition ability. To alleviate this issue, subset selection is often adopted to identify clean samples and filter out noisy ones. Initially, many works focused on designing loss-based subset selection algorithms, \ie, selecting clean samples out of mini-batches based on their losses. To further introduce other usable examples (such as hard examples) into training, \citet{sun2020crssc} proposed a certainty-based reusable sample selection for coping with label noise in web images.

\subsection{Comparative Analysis of Data Sampling and Subset Selection}
Data sampling and subset selection are made in different ways when solving problems. Data sampling methods tend to re-weight the sampling probability or backward losses of all samples in the mini-batch to alleviate the impact of problematic data on model training. In contrast, subset selection methods overcome problems by ignoring some problematic data. Moreover, after the model training, data sampling methods could cover all samples in the training dataset without losing any information, while subset selection methods inevitably lose some data information. Therefore, they have different solution boundaries when dealing with the same problem.

Many data sampling methods have been studied for the class imbalance problem, especially the long-tail distribution problem, but little research has been done on subset selection. This is because subset selection methods usually ignore a large number of samples of the head classes to balance the extremely imbalanced class distribution, which leads to a significant loss of useful information and degradation of model performance on the head classes. Therefore, subset selection is not very good at solving the class imbalance problem compared to data sampling.

\begin{table}[t]
\setlength{\tabcolsep}{2pt}
\small
\caption{Comparative test accuracy (\%) of classical dataset refinement methods on CIFAR-10 and CIFAR-100 \cite{krizhevsky2009learning} datasets with different fractions of noisy\protect\footnotemark. ``Sym.'' indicates symmetric, and ``Asym.'' indicates asymmetric. The symbol ``-'' means the results are unavailable. ``*'' means using additional 1,000 clean images during sampling.}
\begin{tabular}{c|cc|c|cccccc|ccccc}
\hline
\multirow{3}{*}{}                  & \multicolumn{2}{c|}{\multirow{3}{*}{Articles}}                                                                                   & \multirow{3}{*}{Backbone} & \multicolumn{6}{c|}{CIFAR-10}                                      & \multicolumn{5}{c}{CIFAR-100}    \\ \cline{5-15} 
                                   & \multicolumn{2}{c|}{}                                                                                &                           & \multicolumn{5}{c|}{Sym.}                        & Asym. & \multicolumn{5}{c}{Sym.}    \\ \cline{5-15} 
                                   & \multicolumn{2}{c|}{}                                                                              &                           & 20\% & 40\% & 50\% & 80\% & \multicolumn{1}{c|}{90\%} & 40\%       & 20\% & 40\% & 50\% & 80\% & 90\% \\ \hline
\multirow{4}{*}{\begin{tabular}[c]{@{}c@{}}Data \\ Sampling\end{tabular}}     & SPL~\cite{kumar2010self}                                               & NIPS2010       & Resnet-101                & 89.0   & 85.0   & -    & 28.0   & \multicolumn{1}{c|}{-}    & -          & 70.0   & 55.0   & -    & 13.0   & -    \\
                                   & ActiveBias~\cite{chang2017active}     & NIPS2017       & DenseNet                  & -    & 84.3 & -    & -    & \multicolumn{1}{c|}{-}    & -          & -    & 57.8 & -    & -    & -    \\
                                   & reweight~\cite{ren2018learning}*                                      & ICML2018       & WideResNet-28-10          &-      & 86.9 &-      &-      & \multicolumn{1}{c|}{-}     &  -          & -     & 61.3 &  -    & -     & -     \\
                                   & DualGraph~\cite{zhang2021dualgraph}                              & CVPR2021       & ResNet-12                 & 96.7 & -    & 92.2 & 77.2 & \multicolumn{1}{c|}{-}    & 94.1       & 88.7 & -    & 75.8 & 50.2 & -    \\ \hline
\multirow{14}{*}{\begin{tabular}[c]{@{}c@{}}Subset \\ Selection\end{tabular}} & Decoupling~\cite{malach2017decoupling}                                             & NIPS2017       & ResNet-18                 & 87.4 & 83.3 & -    & 36.0   & \multicolumn{1}{c|}{-}    & -          & 57.8 & 49.9 & -    & 17.0   & -    \\
                                   & MentorNet~\cite{jiang2018mentornet} & ICML2018       & Resnet-101                & 92.0   & 89.0   & -    & 49.0   & \multicolumn{1}{c|}{-}    & -          & 73.0   & 68.0   & -    & 35.0   & -    \\
                                   & Co-teaching~\cite{han2018co}             & NIPS2018       & ResNet-101                & 87.3 & 82.8 & -    & 26.2 & \multicolumn{1}{c|}{-}    & -           & 64.4 & 57.4 & -    & 15.2 &  -    \\
                                   & Co-teaching+~\cite{yu2019does}                          & ICML2019       & ResNet-18                 & 88.2 & -    & 84.1 & 45.5 & \multicolumn{1}{c|}{30.1} & -          & 64.1 & -     & 45.3 & 15.5 & 8.8  \\
                                   & INCV~\cite{chen2019understanding}                  & ICML2019       & ResNet-32                 & 89.5 & 86.8 & -    & 53.3 & \multicolumn{1}{c|}{-}    & -          & 58.6 & 55.4 & -    & 23.7 & -    \\
                                   & SELFIE~\cite{song2019selfie}                                & ICML2019       & DenseNet                  & -    & 86.5 & -    & -    & \multicolumn{1}{c|}{-}    & -          & -    & 62.9 & -    & -    & -    \\
                                   & O2U-net~\cite{huang2019o2u}                    & ICCV2019       & ResNet-101                & 92.6 & 90.3 & -    & 37.8 & \multicolumn{1}{c|}{-}    & -          & 74.1 & 69.2 & -    & 39.4 & -     \\
                                   & CL~\cite{lyu2019curriculum}                & ICLR2020       & DenseNet                  & 84.3 & -    & 77.7 & -    & \multicolumn{1}{c|}{-}    & -          & -    & -    & -    & -    & -    \\
                                   & AUM~\cite{pleiss2020identifying}                          & NIPS2020       & ResNet-32                 & 90.2 & 87.5 & -    & 54.4 & \multicolumn{1}{c|}{-}    & -          & 65.5 & 61.3 & -    & 31.7 & -     \\
                                   & TopoFilter~\cite{wu2020topological}                                           & NIPS2020       & ResNet-18                 & 90.2 & 87.2 & -    & 45.7 & \multicolumn{1}{c|}{-}    & -          & 65.6 & 62.0   & -    & 20.7 & -    \\
                                   & CRUST~\cite{mirzasoleiman2020coresets_nips}                         & NIPS2020       & ResNet-32                 & 91.1 & -    & 86.3 & 58.3 & \multicolumn{1}{c|}{-}    & 88.8       & 65.2 & -    & 56.4 & -    & -    \\
                                   & CL: PBNR~\cite{northcutt2021confident}                                 & JAIR2021       & ResNet-50                 & 90.7 & 87.1 & -    & -    & \multicolumn{1}{c|}{-}    & -          & -    & -    & -    & -    & -    \\
                                   & MORPH~\cite{song2021robust}                                & SIGKDD2021 & WideResNet-16-8           & -    & 92.0   & -    & -    & \multicolumn{1}{c|}{-}    & 93.7       & -    & 72.6 & -    & -    & -    \\
                                   & NGC~\cite{wu2021ngc}                            & ICCV2021       & ResNet-18                 & 95.9 & -    & 94.5 & 91.6 & \multicolumn{1}{c|}{80.5} & 90.6       & 79.3 & -    & 75.9 & 62.7 & 29.8 \\
                                   & UNICON~\cite{karim2022unicon}                            & CVPR2022       & ResNet-18                 & 96.0 & -    & 95.6 & 93.9 & \multicolumn{1}{c|}{90.8} & 94.1       & 78.9 & -    & 77.6 & 63.9 & 44.8 \\ \hline
\end{tabular}
\label{ac_noisy}
\end{table}

\footnotetext{We present here only the results reported in published articles due to the failure to reproduce some of the competitive results reported in the articles. This results in the presence of some missing values. We believe that these missing values do not affect the reader's judgment of the best method. Table~\ref{ac_efficient} is the same case.}

In dealing with the problem of noisy labels, the data sampling methods down-weight the sampling probability or losses of mislabeled samples to alleviate the impact of those noisy data on model training. During the training, this family of methods can cover all samples in the training dataset without breaking the distribution of the dataset, but still inevitably accumulate errors from mislabeled samples. Therefore, they work well on moderately noisy data but fail to handle heavily noisy data. As shown in Table~\ref{ac_noisy}, the data sampling method \cite{zhang2021dualgraph}, for example, performs well on the CIFAR-10 dataset with 20\% and 50\% symmetric noise, but the test accuracy drops significantly to 77.2\% on the CIFAR-10 dataset with 80\% high noise. Compared with data sampling, subset selection methods overcome the problem of noisy labels by ignoring all unclean samples that contain many mislabeled instances. Although this family of methods lost some information about the training dataset, it turns out that training on the clean samples yields much better performance than correcting the whole sample on heavily noisy data \cite{han2018co}. Also, as shown in Table~\ref{ac_noisy}, the subset selection method \cite{wu2021ngc} not only performs well on the CIFAR-10 dataset that contains 20\% and 40\% symmetric noise but also still achieves more than 90\% good performance on the CIFAR-10 dataset with 80\% symmetric noise. Therefore, the subset selection method is a more promising approach for heavy noise.

\section{Dataset Refinement for Fair Learning}
\label{sec:DO_FL}
Dataset biases can lead to an overestimation of the true capabilities of current AI systems, so it is crucial for responsible AI to go for fairness against biased data. \citet{sagawa2020investigation} have advocated resampling the majority group rather than weighting the minority group to achieve low worst-case group errors. Therefore, the key to dealing with this problem is to prevent models from overfitting the training dataset or to remove data that could easily allow the model to take shortcuts to fit the training set. This is also the way to improve the generality of the model.

\subsection{Subset Selection}
Dataset bias has various forms, such as contextual bias \cite{torralba2003contextual} in detection and recognition datasets, shape bias \cite{ritter2017cognitive} in ImageNet datasets, and static bias \cite{li2018resound} in action recognition datasets. Different measurers are proposed to help the selector measure these biased data and filter them out. RESOUND \cite{li2018resound} was proposed to quantify the representation biases of action recognition datasets and assemble a new $K$-class dataset with smaller biases by selecting an existing $C$-class dataset $(C > K)$. REPAIR \cite{li2019repair} was designed to learn a probability distribution over the dataset that favors instances that are hard for a given representation and removes representation bias by dataset selection. Some simple and classical subset selection strategies were used, such as Uniform (Keep p = 50\% examples uniformly at random) and Threshold (Keep all examples $i$ such that $w_{i} \geq t$, where t = 0.5 is the threshold). AFLITE \cite{le2020adversarial} was used to measure the ``predictability'' of each image, \ie the ability of simple linear classifiers to predict them correctly, and images with the highest predictability scores were filtered out.

\section{Dataset Refinement for Data-efficient Learning}
\label{sec:DO_DE}

\subsection{Data Sampling}
Stochastic optimizers (\eg Stochastic gradient descent (SGD), mini-batch SGD) are commonly used for model parameter optimization. Instead of computing the full gradient $\bigtriangledown R(f_\theta)$ on the full dataset, they randomly select $n$ samples, compute $\sum_{i}^{n}\bigtriangledown R(f_\theta, x_{i})$ to approximate $\bigtriangledown R(f_\theta)$ and update the model parameters in each iteration. However, the convergence rate is slow when trained on randomly selected samples. Data sampling methods can accelerate convergence rates by carefully adjusting the order of training samples to achieve data-efficient learning. Generally, they re-weight each data according to the designed metric and then select training samples based on the weights before each iteration. Different sampling strategies have been proposed to accelerate the convergence rate well.

Compared to the random sampling of the overall dataset, stratified sampling \cite{zhao2014accelerating} and typicality sampling \cite{peng2019accelerating} are proposed to sample from local data randomly. To be specific, stratified sampling first divides the training set into several groups using a clustering algorithm and then randomly selects data from each group respectively. Typicality sampling first divides the training set into a high-density subset $H$ with size $N_{1}$ and a low-density subset $L$ with size $N_{2}$, then randomly selects $n_{1}$ data from subset $H$ and $n_{2}$ data from subset $L$ with the constraint $\frac{n_{1}}{N_{1}} \geq \frac{n_{2}}{N_{2}}$. The above methods are roughly designed, where the weights of those data in each local subset are still the same.

Another type of method aims to design metrics that measure the effectiveness of each data and calculate the weight of each data individually. There are various metrics that can be used to re-weight the training data to speed up convergence, such as loss \cite{kawaguchi2020ordered, jiang2019accelerating, loshchilov2015online}, gradient norms and bounds \cite{alain2015variance, lee2019meta}, data heterogeneity \cite{lu2021variance}, and Lipschitz constants \cite{schmidt2017minimizing}. To select the most informative and moderately difficult data that can accelerate training, \citet{alain2015variance} proposed using the magnitude of each sample gradient to measure importance. \citet{loshchilov2015online} argue that the data with the greatest losses can accelerate training and that these data should be selected more frequently, so they proposed to utilize the latest known loss value of each data to rank and select them. \citet{kawaguchi2020ordered} showed theoretically that their proposed method could converge at a sublinear rate, where the gradient estimator tends to have higher loss data. 

Apart from using instantaneous metrics to measure importance, training dynamics of DNNs on individual samples are also useful. \citet{zhou2020curriculum} introduced the dynamic instance hardness (DIH) metric and designed a learning strategy called DIH guided curriculum learning (DIHCL), which mainly selects samples with higher DIH. It showed that DIH is a more stable measure than instantaneous hardness. 

\subsection{Subset Selection}

\textbf{Empirical Subset Selection.} Subset selection methods for data-efficient learning have recently gained much interest. Extensive works \cite{lapedriza2013all, wang2018data, swayamdipta2020dataset} have empirically demonstrated that deep learning models may not favor all training samples, and performance and generalization accuracy can be further improved by dropping some unfavorable samples, promoting efficient model learning. Here, the key issue in subset selection is how to determine the effectiveness of per data to select the most informative subset. 

First, \citet{wang2019e2} proposed a straightforward, simple, but surprisingly effective stochastic mini-batch dropping (SMD) method, which skips each mini-batch with a default probability of 0.5. This method of dropping samples randomly with a certain probability has unstable performance, and it may blindly drop highly informative samples, resulting in degraded performance.

\citet{shrivastava2016training} found that the dataset contains a large number of easy examples and a small number of difficult examples through research, and automatically selecting these difficult examples can make training more effective and efficient. Based on this finding, some methods \cite{bachem2015coresets, paul2021deep} use the loss or gradient of each sample to measure and select the difficult subset. Intuitively, data with higher losses are more difficult for model training. However, the instantaneous loss is not a stable measure for subset selection, and training dynamic is commonly used. \citet{toneva2018empirical} introduced the metric \textit{forgetting event} and counted the number of forgetting events that happened during the training. If a forgetting event occurs for a sample, it indicates that the state of this sample has shifted from being correctly predicted to being incorrectly predicted. After the model training, the training data are sorted according to the number of forgetting events, and the data with more forgetting events are selected. The number of forgetting events is a stable metric that reveals the intrinsic properties of the training data. In addition to hand-designed metrics, \citet{fan2017learning} used reinforcement learning to train a neural network that learned to select samples automatically for another neural network to optimize convergence speed.

The above methods are useful tools for machine learning on large datasets. However, many of them are prohibitively expensive to apply in deep learning because they require training the original model \cite{birodkar2019semantic, ghorbani2019data, toneva2018empirical} or extra model \cite{fan2017learning} on the full training set to calculate features or some metrics one or many times. To improve this computational efficiency, \citet{coleman2019selection} proposed performing data subset selection by using a small proxy model, built by removing hidden layers from the target model, using smaller architectures, and training for fewer epochs.

Besides, the above methods do not provide any theoretical guarantee for the quality of the trained model on the extracted subsets and may perform poorly in practice. For this reason, a great deal of work looks to coreset algorithms \cite{feldman2020core} and submodular functions \cite{nemhauser1978analysis, fujishige2005submodular} for subset selection.

\begin{figure}[t]
\begin{center}
\includegraphics[width=\textwidth]{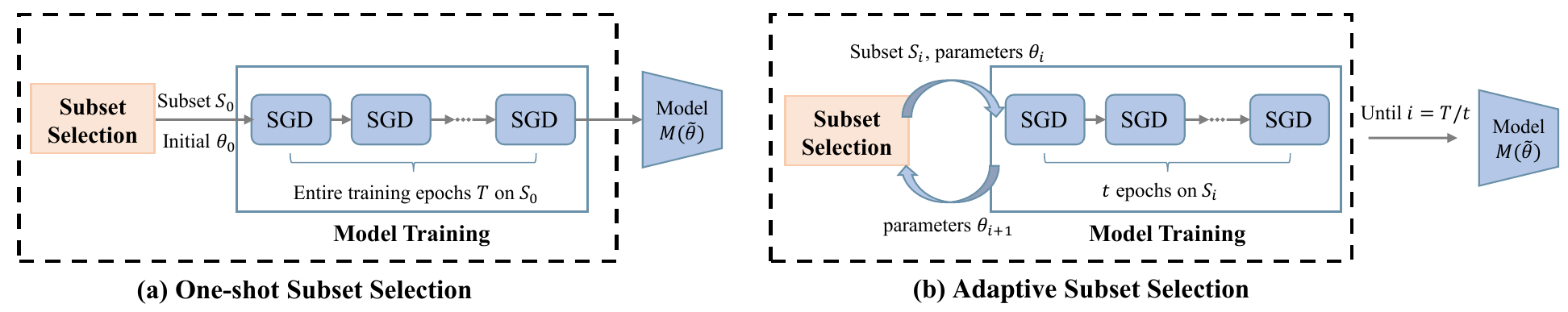}
\end{center}
\caption{Comparison of (a) one-shot subset selection, and (b) adaptive subset selection \cite{haussmann2019deep} pipelines. If a model needs to be trained for $T$ epochs, when using one-shot subset selection methods~\cite{zheng2022coverage}, subset selection is performed only once; when using adaptive subset selection methods~\cite{killamsetty2021grad}, subset selection is performed every $t$ epochs until it is performed $T/t$ times.}
\label{fig:ss}
\end{figure}

\par \textbf{Coreset Selection.} Coresets are weighted subsets of the data that approximate certain desirable characteristics of the full data (\eg the loss function, full gradient, \etc.) \cite{feldman2020core, mirzasoleiman2020coresets}. Early coreset selection algorithms were designed to accelerate the learning and clustering of machine algorithms, such as k-means and k-medians \cite{har2007smaller}, SVM \cite{tsang2005core}, Bayesian logistic regression \cite{huggins2016coresets}, and Bayesian inference \cite{campbell2018bayesian}. However, these algorithms have limited applications for the deep learning models that dominate the field of computer vision because they are designed for specific models and problems. Recently, deep coreset selection-based methods \cite{mirzasoleiman2020coresets, killamsetty2021glister, killamsetty2021grad, killamsetty2021retrieve, killamsetty2022automata} have been proposed, showing great promise for efficient and robust training of deep models.

In the supervised learning scenarios, data subset selection methods have access to the labels of the training dataset. With the supervision of labels, CRAIG \cite{mirzasoleiman2020coresets} selected representative coresets of the training data that closely estimate the full training gradient, \ie sum of the gradients over all the training samples. It has shown promise for several machine learning (ML) models. However, for high dimensional and non-convex ML models, subsets selected based on gradient information only capture gradients along the sharp dimensions and lack diversity within groups of examples with similar training dynamics. Then, ADACORE \cite{pooladzandi2022adaptive} incorporated the geometry of the data to iteratively select coresets of training examples that capture the gradients of the loss preconditioned with the Hessian. Such subsets capture the curvature of the loss landscape along different dimensions and overcome the above issue. RETRIEVE \cite{killamsetty2021retrieve} selected unlabeled data subsets via bi-level optimization for efficient semi-supervised learning. Another approach, GRAD-MATCH \cite{killamsetty2021grad} selected subsets that approximately match the full training loss or validation loss gradient using orthogonal matching pursuit. For faster hyper-parameter tuning, AUTOMATA \cite{killamsetty2022automata} framework was proposed to combine intelligent gradient-based subset selection with hyper-parameter search and scheduling algorithms to achieve it. Analyzed in terms of subset selection pipeline, RETRIEVE~\cite{killamsetty2021retrieve}, GRAD-MATCH \cite{killamsetty2021grad}, and AUTOMATA \cite{killamsetty2022automata} are methods designed based on adaptive subset selection pipeline. These methods differ from the conventional one-shot subset selection methods in that their data selection is performed together with training, and their subsets are gradually refined as the learning algorithm proceeds. The specific pipeline is shown in Fig.~\ref{fig:ss}. In addition, coreset selection is a discrete subset selection problem that is usually solved using a greedy search algorithm~\cite{nemhauser1978analysis, elenberg2018restricted}. When the coreset becomes large, the computational cost becomes high, and the algorithm often produces sub-optimal results. To avoid the trouble of implicit differentiation, \citet{zhou2022probabilistic} proposed a continuous probabilistic bilevel formulation of coreset selection and an efficient solver to the bilevel optimization problem via the unbiased policy gradient.

Besides focusing on the basic goal of efficient model training, several recent articles have further investigated robust learning. CRUST \cite{mirzasoleiman2020noisy} was proposed to handle noise in training data by extending CRAIG \cite{mirzasoleiman2020coresets}. Unlike the conventional minimization on the loss of the training set, GLISTER \cite{killamsetty2021glister} posed the coreset selection problem as a discrete-continuous bilevel optimization problem that minimized the validation set loss for efficient learning focused on generalization.

Apart from the above learning scenarios, coreset selection methods \cite{wei2015submodularity, sener2017active, ash2019deep, killamsetty2021glister} are also used for active learning scenarios, where a subset of data instances from the unlabeled set is selected to be labeled to reduce the labeling cost. Section~\ref{sec:DO_LE} discusses this in detail.

\par \textbf{Subset Selection with Submodular Functions.} In this section, the data subset selection problem is formulated as a constrained submodular optimization problem and addressed using submodular functions as \textit{proxy} functions \cite{wei2015submodularity, kaushal2019learning}. 

Submodular functions \cite{nemhauser1978analysis, fujishige2005submodular}, a special class of set functions, exhibit a property that intuitively formalizes the idea of ``diminishing returns''. Let $V={1, 2, 3, \ldots, n}$ denote a ground set of items (for example, in computer vision, a set of the image or video frame data points in the training set), and $S \subseteq T \subseteq V$. A submodular function $f: 2^{V}\rightarrow \textbf{R}$ will satisfy the diminishing returns property: $f(j \mid S) \triangleq f(S \cup j)-f(S) \geq f(j \mid T)$, where $j \in V \setminus T$. Submodular functions can naturally model the notion of diversity, representation and coverage and are also very appealing because a simple greedy algorithm achieves a $1 - 1/e$ constant factor approximation guarantee \cite{nemhauser1978analysis} for the problem of maximizing a submodular function subject to a cardinality constraint. Therefore, under the constraint of resources (for example, computation resource and restore resource), several recent subset selection works have adopted submodular functions as \textit{proxy} functions \cite{wei2015submodularity, kaushal2019learning, kothawade2022prism} for finding diverse and representative subsets.

Subset selection works well for continuous learning. Continual learning with neural networks has received increasing interest recently. It refers to the setting where a learning algorithm is applied to a sequence of tasks or datasets without the possibility of revisiting old tasks or datasets, usually suffering from the issue of catastrophic forgetting. To prevent forgetting, a replay buffer is usually employed to store the previous data for the purpose of rehearsal. Generally, it stores the random-sampled data as a proxy set, which limits its practicality to real-world applications due to ignoring the fact that not all training data are created equal. Data subset selection is an effective way to deal with this issue. For example, reservoir sampling has been employed in \cite{isele2018selective} so that the data distribution in the replay buffer follows the data distribution that has already been seen. Coverage maximization is also proposed in \cite{isele2018selective}. It uses Euclidean distance as a difference measure to keep diverse samples in the replay buffer. \citet{aljundi2019gradient} proposed a sampling method called GSS that diversifies the samples in the replay memory using the parameter gradient as the feature. Besides, the coreset selection is suitable for it due to the property that a model trained on the coreset performs almost as well as those trained on the full dataset. However, existing coreset selection algorithms are limited to simple models such as k-means and logistic regression. \citet{borsos2020coresets} proposed a novel coreset construction via cardinality-constrained bilevel optimization to efficiently generate coresets for continual learning with neural networks. \citet{yoon2021online} proposed Online Coreset Selection (OCS) to obtain a representative and diverse subset that had a high affinity to the previous tasks from each minibatch during continual learning, and three gradient-based selection criteria were designed according to minibatch similarity, sample diversity and coreset affinity.

\begin{table}[t]
\centering
\setlength{\tabcolsep}{2pt}
\small
\caption{Comparative test accuracy (\%) of some dataset refinement methods for data-efficient supervised learning on balanced and imbalanced datasets (CIFAR-10 and CIFAR-100) with different budgets. Data sampling methods, random method, and Forgetting Events method (subset selection) were tested on original datasets, and other subset selection methods were tested on imbalanced datasets. Note that SB requires additional computation because of the forward passes. The symbol ``-'' indicates that the results are not available.}
\begin{tabular}{ccccccccccccc|c}
\hline
\multicolumn{3}{c}{Dataset:}                                                                                                                                                            & \multicolumn{5}{c}{CIFAR-10}      & \multicolumn{5}{c|}{CIFAR-100}    & \multirow{2}{*}{Backbone} \\ \cline{1-13}
\multicolumn{3}{c}{Budget:}                                                                                                                                                             & 10\% & 20\% & 30\% & 50\% & 100\% & 10\% & 20\% & 30\% & 50\% & 100\% &                           \\ \hline
\multicolumn{1}{c|}{\multirow{6}{*}{Data Sampling}}    & p-SGD\cite{chang2017active}                            & \multicolumn{1}{c|}{NIPS2017}    & -    & 92.3 & 92.9 & 93.9 & 94.6  & -    & 70.2 & 72.1 & 72.9 & 74.6  & ResNet-18                 \\
\multicolumn{1}{c|}{}                                  & c-SGD\cite{chang2017active}                            & \multicolumn{1}{c|}{NIPS2017}    & -    & 91.7 & 92.8 & 93.7 & 94.6  & -    & 69.9 & 71.6 & 73.0   & 74.6  & ResNet-18                 \\ \multicolumn{1}{c|}{}                                  & SB\cite{jiang2019accelerating}                                & \multicolumn{1}{c|}{arxiv2019}   & -    & 93.4 & 93.9 & 94.2 & 94.6  & -    & 70.9 & 72.3 & 73.4 & 74.6  & ResNet-18                 \\
\multicolumn{1}{c|}{}                                  & OSGD\cite{kawaguchi2020ordered}        & \multicolumn{1}{c|}{AISTATS2020} & -    & 90.6 & 91.8 & 93.5 & 94.6  & -    & 70.1 & 72.2 & 73.4 & 74.6  & ResNet-18                 \\
\multicolumn{1}{c|}{}                                  & scan-SGD\cite{2021_BMVC_importanceSampling}                            & \multicolumn{1}{c|}{BMVC2021}    & -    & 92.0   & 93.1 & 93.8 & 94.6  & -    & 70.9 & 72.3 & 73.5 & 74.6  & ResNet-18                 \\
\multicolumn{1}{c|}{}                                  & unif-SGD\cite{2021_BMVC_importanceSampling}                            & \multicolumn{1}{c|}{BMVC2021}    & -    & 91.8 & 92.7 & 93.7 & 94.6  & -    & 70.4 & 72.0   & 73.4 & 74.6  & ResNet-18                 \\ \hline
\multicolumn{1}{c|}{\multirow{8}{*}{Subset Selection}} & \multicolumn{1}{c}{random}    & \multicolumn{1}{c|}{}                                                                                               & 75.7 & 87.1   & 90.2 & 93.3    & 95.6  & 35.2 & 57.4 & 63.9 & 68.6    & 78.6  & ResNet18                  \\
\multicolumn{1}{c|}{}                                  & Forgetting Events\cite{toneva2018empirical}                & \multicolumn{1}{c|}{ICLR2019}    & 67.0    & 86.6    & 91.7 & 94.1    & 95.6     & 34.0    & 56.1    & 64.5 & 72.7   & 78.6     & ResNet18                 \\
\multicolumn{1}{c|}{}                                  & GRAIG
\cite{mirzasoleiman2020coresets}                             & \multicolumn{1}{c|}{ICML2020}    & 87.5 & 90.8 & 92.5 & -    & 95.1  & 55.2 & 66.2 & 70.0   & -    & 75.4  & ResNet18                  \\
\multicolumn{1}{c|}{}                                  & GRAD-MATCH\cite{killamsetty2021grad} & \multicolumn{1}{c|}{ICML2021}    & 90.9 & 91.7 & 91.9 & -    & 95.1  & 59.9 & 68.3 & 71.5 & -    & 75.4  & ResNet18                  \\
\multicolumn{1}{c|}{}                                  & GRAD-MATCH-WARM\cite{killamsetty2021grad} & \multicolumn{1}{c|}{ICML2021}    & 92.2 & 92.1 & 92.0 & -    & 95.1  & 68.2 & 71.3 & 74.1 & -    & 75.4  & ResNet18                  \\
\multicolumn{1}{c|}{}                                  & GRAD-MATCHPB\cite{killamsetty2021grad} & \multicolumn{1}{c|}{ICML2021}    & 90.0 & 93.3 & 93.8 & -    & 95.1  & 60.4 & 70.9 & 72.6 & -    & 75.4  & ResNet18                  \\
\multicolumn{1}{c|}{}                                  & GRAD-MATCHPB-WARM\cite{killamsetty2021grad} & \multicolumn{1}{c|}{ICML2021}    & 92.3 & 93.6 & 94.2 & -    & 95.1  & 69.6 & 73.2 & 74.6 & -    & 75.4  & ResNet18                  \\
\multicolumn{1}{c|}{}                                  & GLISTER\cite{killamsetty2021glister}       & \multicolumn{1}{c|}{AAAI2021}    & 91.9 & 92.8 & 93.6 & -    & 95.1  & 44.0   & 61.6 & 70.5 & -    & 75.4  & ResNet18                  \\
\hline
\end{tabular}
\label{ac_efficient}
\end{table}

\subsection{Comparative Analysis of Data Sampling and Subset Selection}
Data-efficient learning is necessary when real-world applications have limited training resources. During the optimization of the model, data efficient learning can be achieved by performing data sampling or subset selection with budget constraints. Table~\ref{ac_efficient} shows the test accuracy of some data sampling and subset selection methods under various budget restrictions. If a model is trained for $T$ epochs on a dataset that has $N$ samples without a budget restriction, then when there is a budget constraint $B \in [0, 1]$, the budget $B$ is typically used to limit (1) the number of training epochs to $T \times B$ (in scan-SGD~\cite{2021_BMVC_importanceSampling}), (2) the number of samples per epoch to $N \times B$ (in p-SGD~\cite{chang2017active}, c-SGD~\cite{chang2017active}, SB~\cite{jiang2019accelerating}, OSGD~\cite{kawaguchi2020ordered}, unif-SGD~\cite{2021_BMVC_importanceSampling}, and adaptive subset selection methods \cite{mirzasoleiman2020coresets, killamsetty2021grad, killamsetty2021glister}), (3) the number of samples during entire training epochs to $N \times B$ (in one-shot subset selection methods, such as forgetting events~\cite{toneva2018empirical}). Comparing (2) and (3), the subset selected by the former varies with the training epochs, \ie the model can see all N data throughout the training epochs, while the subset selected by the latter can be fixed, \ie it can see only a selected number of $N \times B$ data. For a more visual illustration, we set $B=0.3$ and sampled the original CIFAR-10 dataset using data sampling
\begin{figure*}[t]
\centering
\subfigure[Full training set]{
\label{fig:datamap_full}
\includegraphics[width=0.32\textwidth]{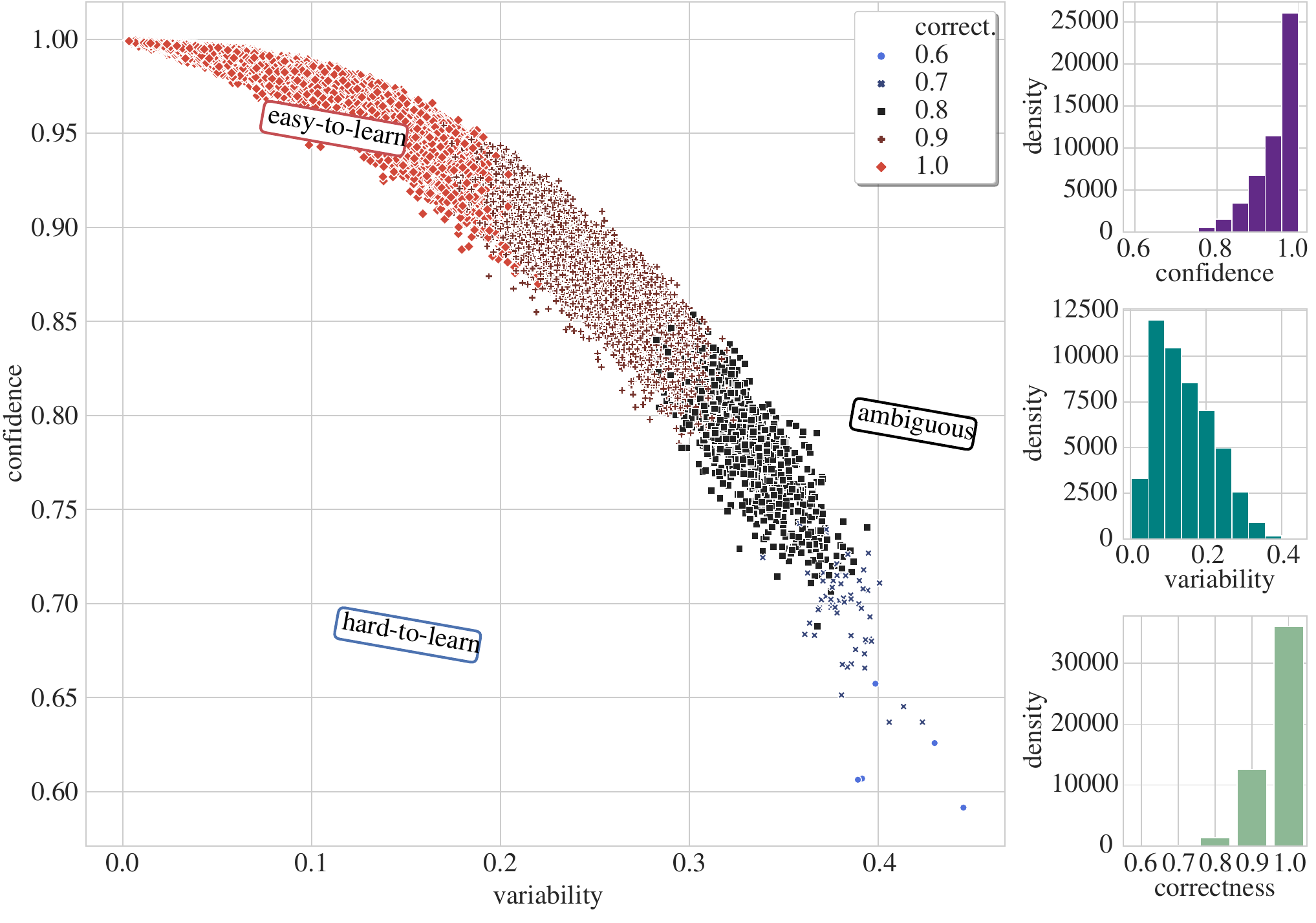}}
\subfigure[Random (one-shot subset selection)]{
\label{fig:random}
\includegraphics[width=0.32\textwidth]{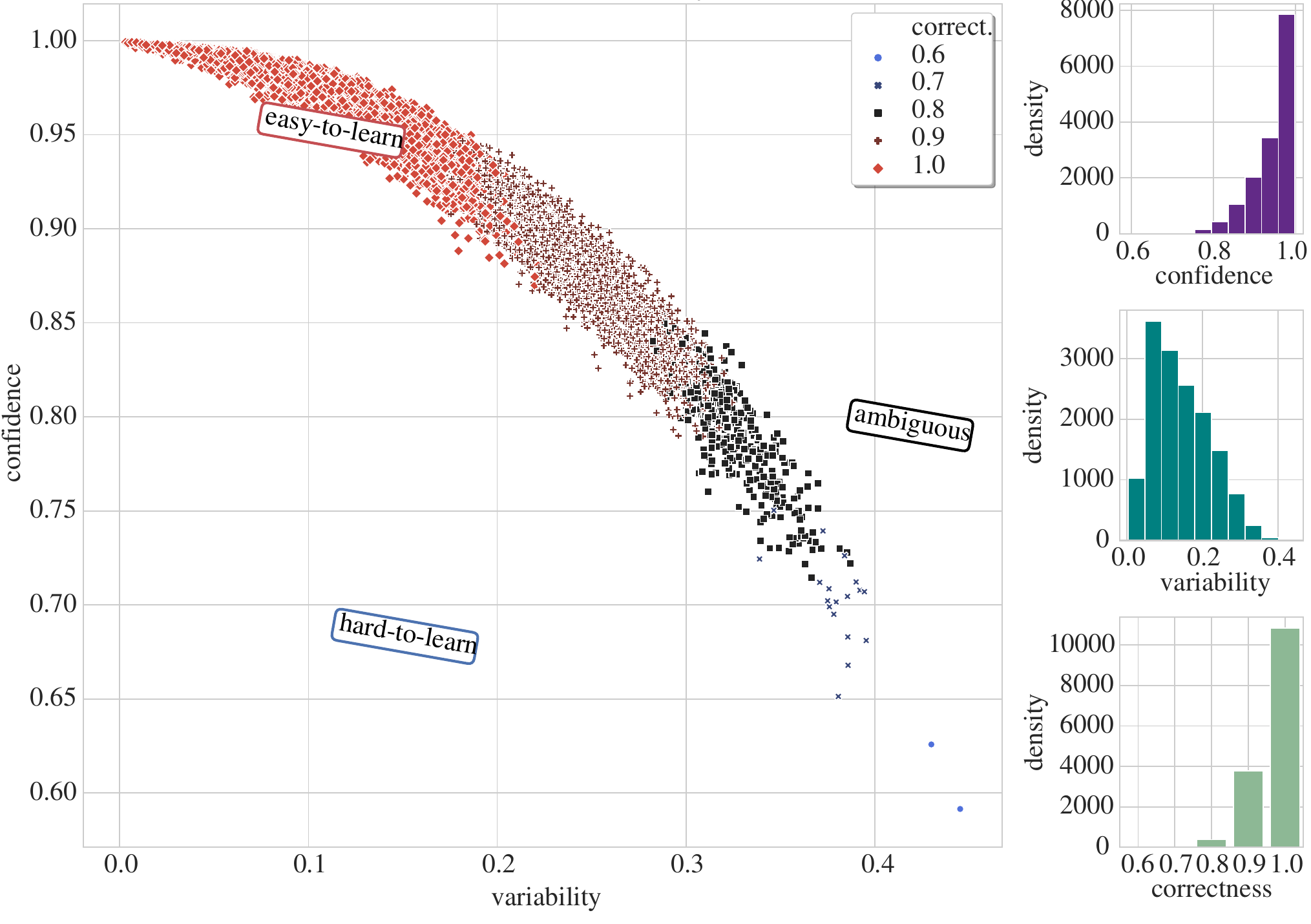}
}
\subfigure[Forgetting Events (one-shot subset selection)]{
\label{fig:forgetting}
\includegraphics[width=0.32\textwidth]{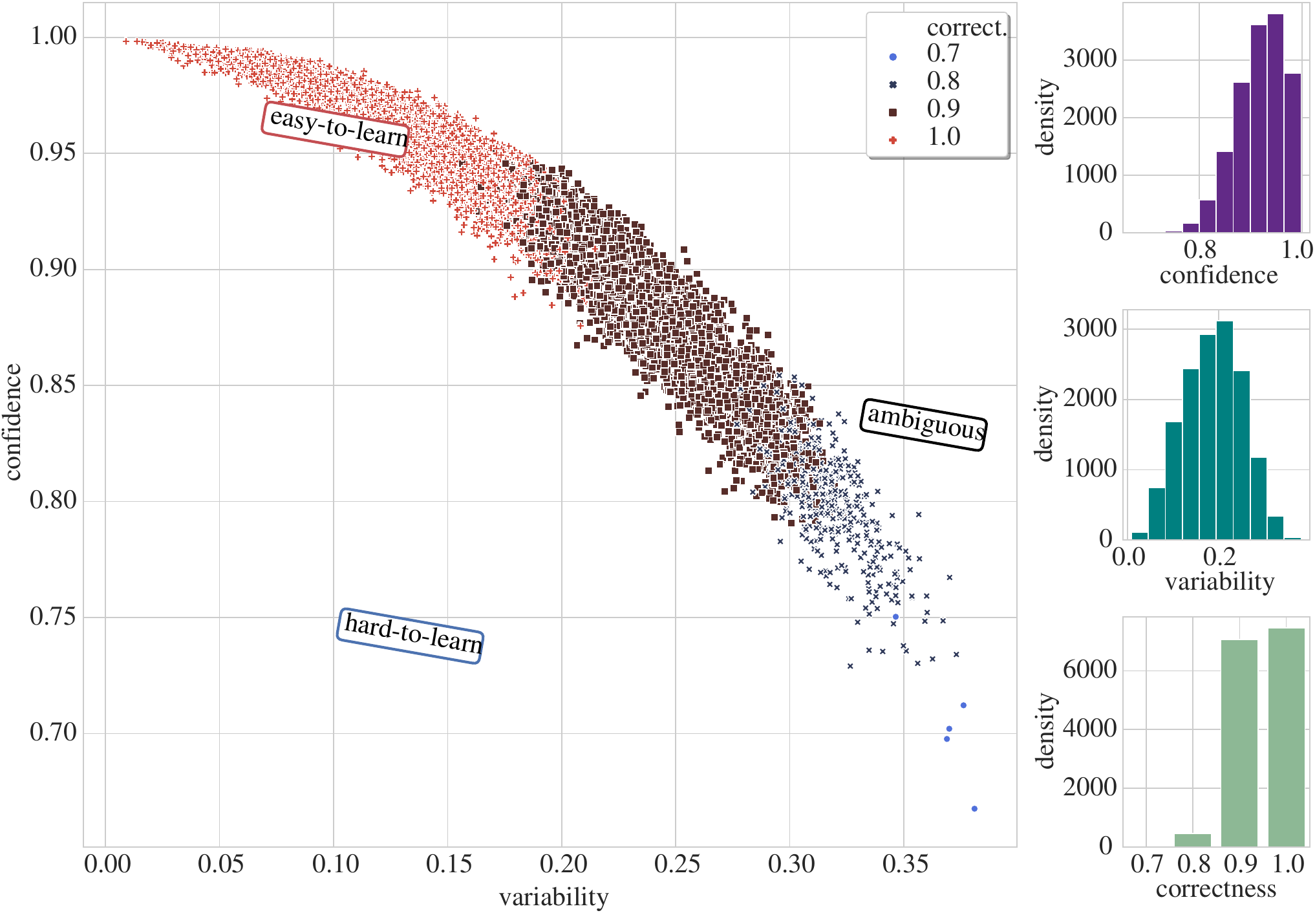}
}
\caption{Data maps for samples sampled by (a) data sampling or dynamic subset selection, (b) random and (c) forgetting events methods.}
\label{fig:data_maps}
\end{figure*}
methods, adaptive sampling methods, the random method (one-shot subset selection), and the forgetting events method (one-shot subset selection); then, we recorded the samples sampled by each method throughout the training periods; finally, the samples for each method are visualized separately in Fig.~\ref{fig:data_maps} with the help of data maps~\cite{swayamdipta2020dataset}, a model-based tool for characterizing and diagnosing datasets. It can be seen that the samples sampled by data sampling and dynamic subset selection cover the full training set, while the samples sampled by the random and forgetting events methods cover 30\% of the full training set. Therefore, one-shot subset selection methods are applicable when the storage budget is limited, such as for the continuous learning task. In addition, the data maps of the samples sampled by the random method are similar to the full training set, while the forgetting events method tends to select the harder (with lower confidence, higher variability, and lower correctness) samples. Table~\ref{ac_efficient} shows the results of some refinement methods for data-efficient supervised learning on CIFAR-10 and CIFAR-100 datasets with different budgets. Data sampling methods, random method, and Forgetting Events method (subset selection) were tested on original datasets (class balance), and other subset selection methods were tested on imbalanced datasets (class imbalance). The imbalanced datasets are formed by removing 90\% of the instances from 30\% of the classes available. The class-imbalanced dataset is obviously smaller and more difficult than the class-balanced dataset. As can be seen in Table~\ref{ac_efficient}, data sampling methods and subset selection methods have comparable results given comparable computation power budgets. Therefore, it can be reasoned that the subset selection methods will perform better than the data sampling methods on the original dataset.

\section{Dataset Refinement for Label-efficient Learning}
\label{sec:DO_LE}
\subsection{Active Learning}
Active learning contributes prominently to label-efficient learning and is attracting increasing attention. Extensive surveys \cite{liu2021survey, ren2021survey} have been conducted on this topic, so we will only cover the most prominent techniques here.

\begin{figure}[t]
\begin{center}
\includegraphics[width=0.65\textwidth]{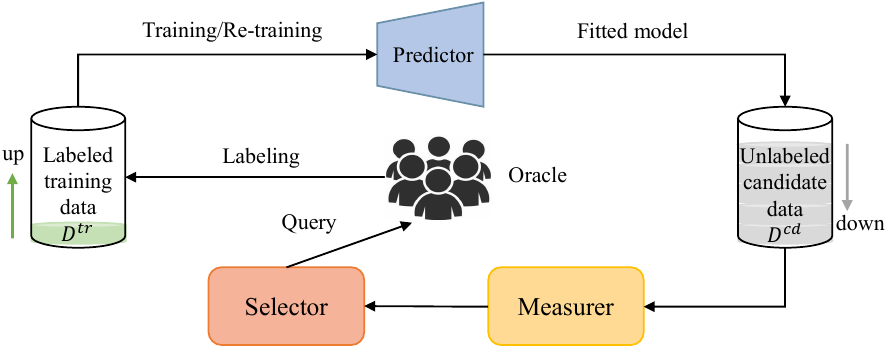}
\end{center}
\caption{The pool-based active learning framework. Initially, active learning trains the predictor with a small amount of labeled training data, uses the designed measurer to measure the importance of the data in the unlabeled candidate data pool, and uses its selector to select a few unlabeled samples to the oracle (\eg human annotator), labels them, adds them into the labeled training data pool, and then re-trains the predictor. During active learning, the amount of data in the unlabeled candidate pool decreases while the amount of data in the labeled pool increases until the label budget is exhausted or the pre-defined termination conditions are reached. Herein, $D^{cl} = D^{tr} \cup D^{cd}$ and $|D^{tr}| < |D^{cl}|$.}
\label{AL}
\end{figure}

Most active learning methods employ pool-based sampling, whose framework is shown in Fig.~\ref{AL}. It selects the data to be annotated based on the evaluation and ranking of the overall unlabeled candidate pool. Similarly, the design of the measurer and the selector is the key to active learning. The main designs include the uncertainty-based methods \cite{lewis1995sequential, seung1992query, beluch2018power, ranganathan2017deep}, distribution-based methods \cite{bilgic2009link, guo2010active, nguyen2004active, sener2017active, sener2017geometric}, and auto-metric learning \cite{contardo2017meta, ravi2018meta, fang2017learning}. 

\textbf{Uncertainty-based methods.} Uncertainty sampling of unlabeled samples is a commonly adopted selection strategy in active learning. It is based on the simple concept of selecting those unlabeled samples that are the most uncertain in terms of a predicted probability for labeling. They assume that the uncertainty samples will provide more information for the training of predictors. There are various ways to measure uncertainty on the predicted probability, such as the common uncertainty-based methods 1-3 in Table~\ref{tab:formula}. For example, when the predictor is a binary classifier, uncertainty sampling selects the example with a probability nearest 0.5. In a multiclass classification task, the least confident metric \cite{ruzicka2020deep} can be used to select the uncertain example. The larger the value of $\phi_\textit{least confidence}({x})$, the more uncertain the data will be. However, when there are confusable samples, the least confident metric will ignore many possible samples. For this problem, the margin sampling ($\phi_\textit{margin}({x})$) \cite{joshi2009multi}, a modified version, was designed to select the example with the smallest probability difference between the most and second most likely labels ($y_{1}$ and $y_2$ in Table~\ref{tab:formula}, respectively). That is, the larger $\phi_\textit{margin}({x})$ is, the higher the uncertainty of the data. This method can be further optimized using entropy as the uncertainty measure ($\phi_\textit{entropy}({x})$) \cite{settles2012active} where entropy is an information-theoretic measure for the amount of information to encode a distribution. Compared to the least confidence and margin confidence, the entropy metric takes into account the predicted probability of the model for all classes of a given data. \textit{Query-by-Committee} \cite{seung1992query} also belongs to the uncertainty-based method. Instead of training only one model, it is done by training a committee of models on the same labeled dataset and selecting samples based on the votes of each model. The more disagreement between models on a given data, the higher the uncertainty of that data.

\textbf{Distribution-based methods.} Simple uncertainty sampling methods focus on individual examples instead of the entire set of examples, resulting in a lack of diversity and representativeness in the selected data. There are three representative data distribution-based methods that can complement uncertainty sampling: diversity, representativeness, and coreset. A diversity-based method can select samples that maximize batch diversity. However, it may select points that are diverse but provide little new information to the model, so most methods \cite{wu2022entropy, elhamifar2013convex, guo2010active, demir2010batch} typically combine diversity and uncertainty for active learning. \citet{demir2010batch} used a heuristic to first filter the pool based on uncertainty and then chose points to label using diversity. \citet{elhamifar2013convex} designed a discrete optimization problem for balancing uncertainty and diversity to obtain a diverse set of hard examples. \citet{wu2022entropy} designed a two-stage approach. An entropy-based non-maximal suppression was proposed to remove instances with redundant information gains in the first stage; in the second stage a diverse prototype strategy was explored to ensure the intra-class diversities and inter-class diversities. Besides diversity, the most commonly used distribution-based method is probably the \textit{representativeness} \cite{elhamifar2015dissimilarity, yang2019single}. Since data is always redundant, it is intuitively good to train models with more representative samples. In \cite{yang2019single}, Yang \etal proposed single-shot active learning using pseudo annotators, where the pseudo annotators can be taken as a special way to find the most representative sample. \textit{Coreset} selection can also be used to find the representative samples for active learning whose representation comes from the global training set, not from a subset. Some methods \cite{wei2015submodularity, killamsetty2021glister, ash2019deep} simplify the active learning problem by reducing it to a submodular optimization that can model characteristics such as diversity and representation. BADGE \cite{ash2019deep} samples groups of points with a diverse and higher magnitude of the hypothesized gradient, incorporating both prediction uncertainty and sample diversity into each selected batch. \citet{sener2017active} defines active learning as a core set selection problem and has shown that models learned over k-centers of a dataset are competitive with those trained over the full data. In this method, selecting center points means that the largest distance between a data repeat point and its nearest center is minimized. Coreset methods have shown promising performances in some datasets. 

In addition to the above approach, feature representation also can be used to select data. \citet{parvaneh2022active} proposed an Active Learning by FeAture Mixing (ALFA-Mix) method that identifies unlabelled instances with sufficiently distinct features by looking for prediction inconsistencies resulting from interventions on their representations. This method has advantages for large tasks with a large number of classes and in low data regimes. 

\begin{figure}[t]
\begin{center}
\includegraphics[width=\textwidth]{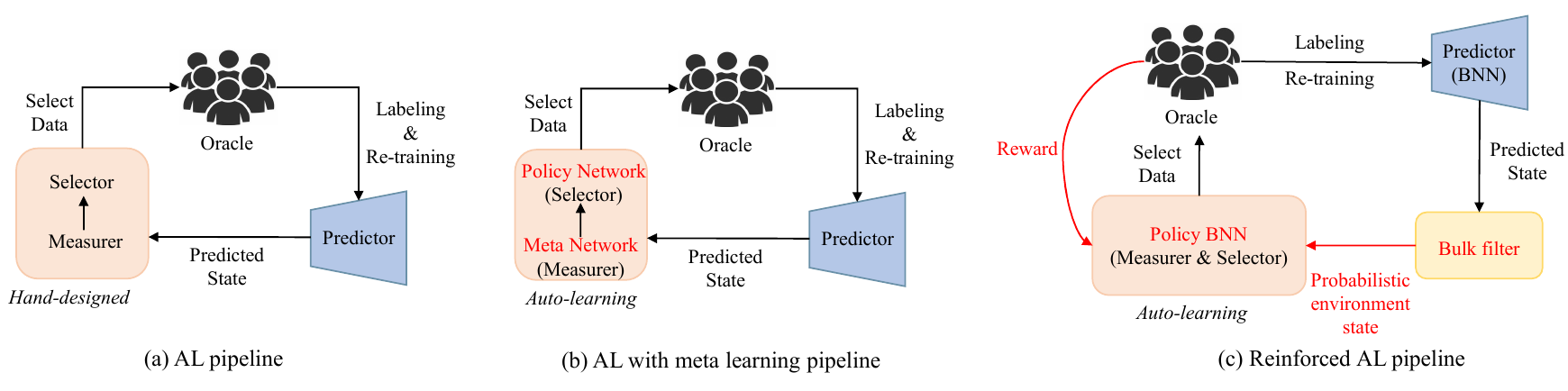}
\end{center}
\caption{Comparison of (a) standard AL, (b) AL with meta learning, and (c) Reinforced AL \cite{haussmann2019deep} pipelines. Different from (a), the measurer and selector of (b) and (c) are equipped with a well-designed deep network structure and will automatically learn from the data. The differences are emphasized in red. Notably, to obtain well-learned measures and selectors, the structure of the training set needs to be reformed prior to learning.}
\label{3ALs}
\end{figure}

\textbf{Auto-metric learning methods.} The metrics used in the above methods are manually designed based on human experience and prior assumptions. Thus, the performances of those methods are severely limited by the experience of the researcher. In contrast to them, there is a lot of work focused on auto-metric learning. They are trying to design measurers with deep architectures that automatically learn how to measure the importance of each data but may need more data during the training. Different learning modes are introduced to active learning, including meta-learning and reinforcement learning, as shown in Fig.~\ref{3ALs} (b) and (c). Meta learning, which aims to obtain higher training performances with fewer training samples, is introduced to form the meta learning measurer \cite{ravi2017optimization, contardo2017meta, ravi2018meta}. The meta learning measurer outputs the importance of the unlabeled samples but not the prediction error. After ranking the importance of unlabeled samples, the top $n$ samples are usually selected and labeled. Reinforcement learning is introduced in \cite{fang2017learning} to flexibly and quickly tune the selection policy based on external feedback. The comparison of standard AL, AL with meta learning, and reinforced AL pipelines are shown in Fig.~\ref{3ALs}. Unlike the traditional hand-designed policies of AL, measurers and selectors in AL with meta learning pipeline and reinforced AL pipeline have well-designed deep network structures. This shifts the spotlight from how to design an effective manual strategy to how to design an effective network structure. While network structures are generally rich and complex, they can effectively take into account the uncertainty and diversity of data during automatic learning. Besides, to train these measurers and selectors well, the structure of the training set needs to be reformed before training \cite{liu2021survey}. 

After achieving label-efficient training, reducing test labeling will be the main bottleneck to reducing the overall labeling cost, as the test data also needs to be manually labeled. Especially when the performance of rare categories is of interest, the validation problem is particularly difficult. While it is easy to collect a large amount of unlabeled data, finding even a small number of positives via uniform random sampling can require labeling thousands of images. Therefore, it requires more efficient methods than uniform random sampling. \citet{poms2021low} proposed an active sampling algorithm for simultaneous calibration and importance sampling to achieve accurate estimation even in the case of low samples (< 300 samples).

\subsection{Comparative Analysis of Subset Selection and Active Learning}
In terms of the nature of the task, they both aim to select the most informative subset from the training set, \ie the core set. Thus, coreset selection algorithms \cite{kaushal2019learning, wei2015submodularity} are suitable for both subset selection and active learning. However, there are differences between the setting of the two tasks. The purpose of the data subset selection algorithm is to reduce the amount of data, and the data studied are labeled. Therefore, the subset selection can be supervised with labels, and richer metrics are used to select the core set. In contrast, active learning aims to reduce the number of labels, and the data studied is unlabeled, and the methods mainly focus on exploring the uncertainty of the data to select the core set. Therefore, a core set selection algorithm needs to be fine-tuned when applied to two tasks.

\section{Relevant Learning Problems}
\label{sec:relevant}

\subsection{Dataset Distillation}
Dataset distillation (or dataset condensation) \cite{cazenavette2022dataset, wang2018dataset, zhao2021dataset} encapsulates knowledge of the full training set into a small number of synthetic training images, with the aim that a DNN trained on the synthesized data has almost the same or better generalization ability than one trained on the full training set. In contrast to subset selection and data sampling, dataset distillation is not limited to the sample distribution in the original dataset when synthesizing data points, and it does not pursue the realism of the generated data. In addition, data privacy is protected because dataset distillation does not present the original data to the user. Therefore, we consider dataset distillation as an effective and promising way to address the data privacy problem. Although significant progress has been made in this research area, it remains a challenge to apply dataset distillation on large-scale and high-resolution datasets due to expensive and difficult optimizations.

\subsection{Feature Selection}
In the era of big data, feature selection \cite{li2017feature} is one of the very popular ways of data pre-processing. Unlike the data subset selection that selects data, feature selection selects the features of the data. Generally, when training on a dataset, apart from the presence of redundant and noisy data, there are many irrelevant and noisy features that not only increase the computational cost but also cause the model to overfit and perform poorly on unseen data. Feature selection aims to remove these features while maintaining the physical meaning of the original features to reduce storage and computation costs while avoiding significant losses. Typically, there are three types of feature selection: supervised feature selection, unsupervised feature selection, and semi-supervised feature selection. Thus, feature selection can be complemented with data subset selection and active learning to achieve effective and efficient model training. In addition, from an optimization perspective, feature selection and data subset selection are NP-hard problems that are often solved using greedy algorithms~\cite{nemhauser1978analysis}. However, it has been found through research that in addition to greedy algorithms, metaheuristic algorithms~\cite{dokeroglu2022comprehensive, bacanin2022hybridized, tuba2022acute, franco2022hybrid} are often used to solve the optimization problem of feature selection. Moreover, it has been shown that metaheuristic algorithms perform better than exhaustive or greedy methods~\cite{deniz2017robust}. Therefore, we believe that metaheuristics are also valuable for data selection, which currently lacks relevant research and is yet to be studied.

\subsection{Semi-supervised Learning}
Semi-supervised learning (SSL) algorithms have achieved great success in recent years. It combines limited labeled data with large amounts of unlabeled data to perform certain machine learning tasks. However, the current state-of-the-art SSL algorithms are computationally expensive and require significant computational time and resources. This is a huge limitation for many small companies and academic groups. Since coreset selection and SSL are orthogonal, coreset selection algorithms can be utilized to provide an informative unlabeled subset of data for SSL to facilitate research and effective training of SSL. By training on the unlabeled data subset instead of the entire unlabeled dataset, SSL can achieve faster convergence and significantly reduce computational costs.

\section{Discussion and Future Directions}
\label{sec:future}

\textbf{Green AI.} With the continuous development and maturity of model structures, good model structures are readily available and are no longer the main obstacle, while data becomes the main factor in determining the effectiveness and trustworthiness of AI. Data-centric AI has become an inevitable trend. As one of the key technologies, dataset refinement is receiving more and more attention. However, progress in each dataset refinement direction has been inconsistent. Active learning, for example, has been successful in common computer vision downstream tasks, while data subset selection has been studied mainly in image classification and recognition tasks. Essentially, the research progress achieved is positively correlated with the focus on data problems in the field. This suggests that the field of computer vision is currently generally plagued by problems such as costly data annotation and noisy labels, while the problem of slow computation due to large data scales can be solved by increasing computational resources \cite{10.1145/3225058.3225069}. Inspired by the call for green AI \cite{schwartz2020green}, we argue that not only training performance but also training efficiency needs to be considered. Therefore, dataset refinement for data-efficient learning has remarkable research value and urgently needs to be applied and studied in various computer vision tasks.

\textbf{Dataset refinement for heterogeneous data biases.} In addition, existing dataset refinement studies often implicitly assume that the dataset has only one specific data problem and focus on solving this single problem. Differently, multiple problems often coexist in real datasets, \eg data crawled from the web and data from real applications (autonomous driving and medical diagnosis) can contain both noisy labels, imbalanced class distributions, and data privacy problems. The dataset refinement to solve a single problem is not directly applicable to real datasets with heterogeneous data biases. For example, in the problem of noisy labels, data sampling tends to up-weight the data with smaller training losses as they are more likely to be clean data. Nevertheless, in the problem of class imbalance, the data with smaller training losses are more likely to be the head classes. Undoubtedly, up-weighting samples with smaller losses will exacerbate the problem of class imbalance. To better optimize real datasets and improve dataset quality in real applications, dataset refinement methods need to have the ability to solve multiple data problems simultaneously. Such methods are yet to be investigated. To our knowledge, the meta sampler \cite{shu2022cmw} is a promising solution that adaptively learns the data weights for different datasets.

\textbf{Dataset refinement with theoretical guarantees.} Commonly used dataset refinement methods, such as loss-based, confidence-based, and uncertainty-based methods, are empirically designed methods without theoretical guarantees on the quality of the models trained on the refined dataset. For example, with respect to the different methods for subset selection, it is known from the comparative experimental results of \cite{guo2022deepcore} that the submodular function-based method with theoretical guarantees has an obvious performance advantage. Therefore, how to better design methods with theoretical guarantees is worth further exploration.

\section{Conclusion}
\label{sec:conclusion}
Dataset refinement is a key technique for improving dataset quality and has received increasing attention in machine learning and deep learning. In this article, for the first time, we summarize and analyze inherent or extrinsic problems faced by large-scale datasets. In response to these problems, we review the field of dataset refinement, and present and compare different dataset refinement algorithms in detail. Based on our analysis, we provide several potential future research directions. We hope that this survey will provide useful insights for researchers and motivate them to make more progress in the future.

% \begin{acks}
% This work was supported by National Key R\&D Project (2021YFC3320301), National Natural Science Foundation of China (62171325), Hubei Key R\&D (2022BAA033) and CAAI-Huawei MindSpore Open Fund. The Supercomputing Center of Wuhan University supports the supercomputing resource. The authors would like to thank the editors and the anonymous reviewers for their insightful comments and remarks.
% \end{acks}

\bibliographystyle{ACM-Reference-Format}
\bibliography{sample-base}

\end{document}